# A comprehensive control architecture for semi-autonomous dual-arm robots in agriculture settings

Jozsef Palmieri, Paolo Di Lillo, Stefano Chiaverini, Alessandro Marino

University of Cassino and Southern Lazio

Department of Electrical and Information Engineering

Please cite as:

# A comprehensive control architecture for semi-autonomous dual-arm robots in agriculture settings★

Jozsef Palmieri[a], Paolo Di Lillo[a,*], Stefano Chiaverini[a,*] and Alessandro Marino[a]

[a]*Department of Electrical and Information Engineering*
*University of Cassino and Southern Lazio*
*Via G. Di Biasio 43*

ARTICLE INFO

ABSTRACT

The adoption of mobile robotic platforms in complex environments, such as agricultural settings, requires these systems to exhibit a flexible yet effective architecture that integrates perception and control. In such scenarios, several tasks need to be accomplished simultaneously, ranging from managing robot limits to performing operational tasks and handling human inputs. The purpose of this paper is to present a comprehensive control architecture for achieving complex tasks such as robotized harvesting in vineyards within the framework of the European project CANOPIES. In detail, a 16-DOF dual-arm mobile robot is employed, controlled via a Hierarchical Quadratic Programming (HQP) approach capable of handling both equality and inequality constraints at various priorities to harvest grape bunches selected by the perception system developed within the project. Furthermore, given the complexity of the scenario and the uncertainty in the perception system, which could potentially lead to collisions with the environment, the handling of interaction forces is necessary. Remarkably, this was achieved using the same HQP framework. This feature is further leveraged to enable semi-autonomous operations, allowing a human operator to assist the robotic counterpart in completing harvesting tasks. Finally, the obtained results are validated through extensive testing conducted first in a laboratory environment to prove individual functionalities, then in a real vineyard, encompassing both autonomous and semi-autonomous grape harvesting operations.

## 1. Introduction

The growing need for flexibility in several production scenarios has led to the employment of increasingly complex robotic systems (Ghodsian et al., 2023). This is especially evident in fields such as logistics, where dual-arm robots are often required in picking operations when the objects to be picked have very different shapes (Garabini et al., 2021), ground structure assembly applications (Štibinger et al., 2021), in which mobile manipulators are employed to manage large workspaces, and the inspection and maintenance of high-altitude platforms (Ollero et al., 2022), which require equipping manipulators with flying bases to reach the inspection sites.

Additionally, unlike traditional industrial manufacturing scenarios where robots often perform repetitive tasks in a well-known and fixed environment, recent years have seen robots adopted in unstructured environments for executing very complex manipulation operations. Examples are the maintenance and operation of Oil & Gas underwater structures (Di Lillo et al., 2021b), underground search and rescue operations (Wang et al., 2023), and precision agriculture settings (Palmieri et al., 2024), which is the main scenario tackled in this paper, where robots must autonomously execute challenging tasks such as harvesting and pruning, often in environments with significant perception uncertainty and physical interaction requirements. Working in such scenarios requires the robots to have more complex structures and reasoning capabilities compared to those employed in traditional manufacturing settings. As a simple example, more and more dual-arm systems, potentially mounted on a mobile base, are employed. In this case, several constraints must be managed at the control level to avoid collisions among the different parts of the robot, such as between arms or between arms and the mobile base

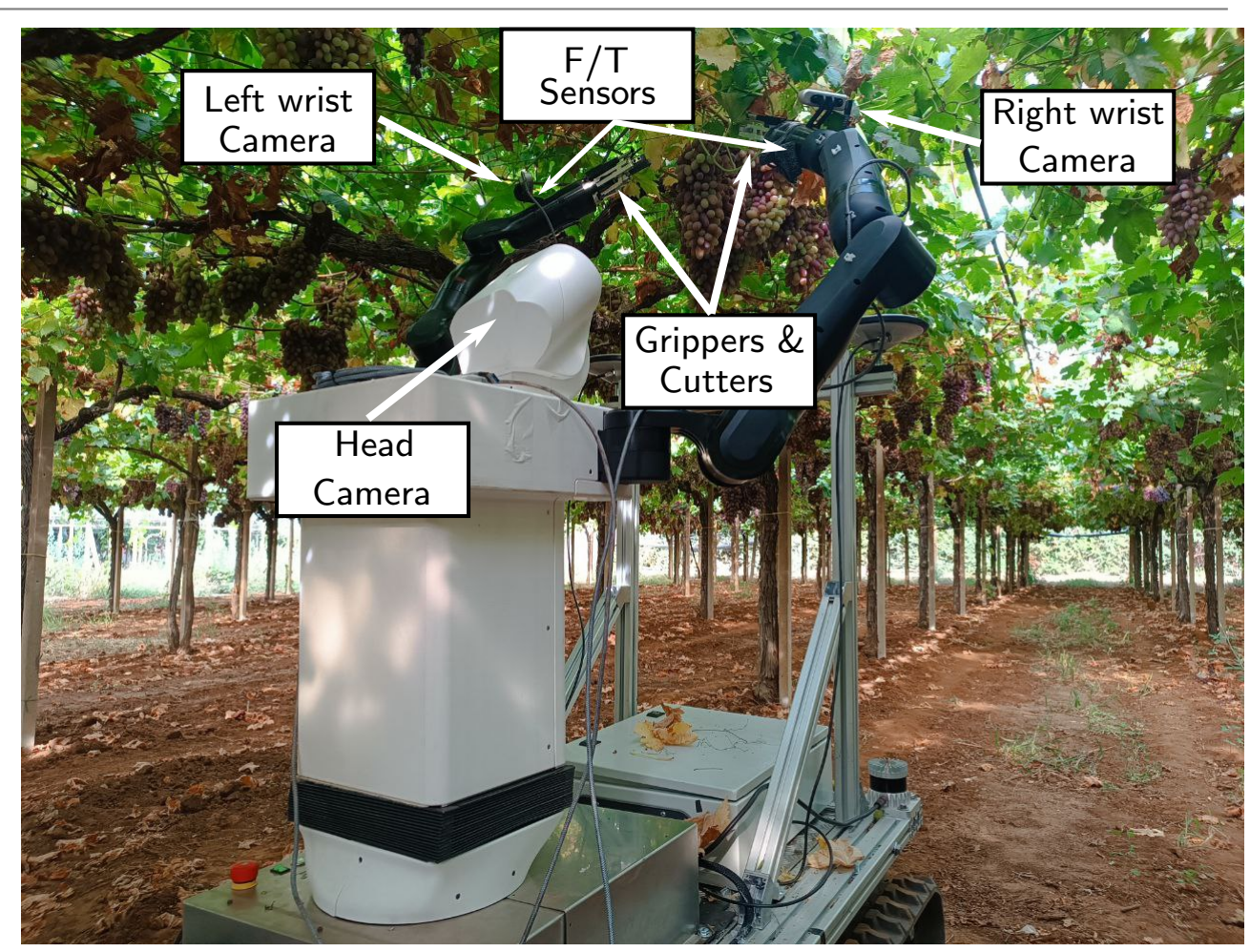

**Figure 1:** Farming robot designed for the EU-funded project CANOPIES. In particular, it highlights: the head and the wrist cameras, the end-effectors equipped with gripper and cutter, and Force/Torque (F/T) sensors.

★Conflict of interest - none declared
*Corresponding author
jozsef.palmieri@unicas.it (J. Palmieri); pa.dilillo@unicas.it (P. Di Lillo); chiaverini@unicas.it (S. Chiaverini); al.marino@unicas.it (A. Marino)
ORCID(s): 0009-0000-5863-5832 (J. Palmieri); 0000-0003-2083-1883 (P. Di Lillo); 0000-0002-7891-0122 (S. Chiaverini); 0000-0002-9050-9825 (A. Marino)

itself. The challenge is to properly coordinate the motion of the different parts effectively, designing a control algorithm built on a model of the system as a whole. Moreover, these robotic systems usually have additional Degrees of Freedom (DOFs) that can be exploited to generate internal motions aimed at optimizing the robot's posture. Furthermore, the complexity of these systems further increases both in terms of sensor suites and control architectures considering the unstructured nature of the environments they are required to work in, especially when they have to collaborate with human operators, possibly enabling physical human-robot interaction (pHRI) paradigms that further enhance robot effectiveness and efficiency (Selvaggio et al., 2021). In these settings, robots must not only plan and execute complex movements but also dynamically interact with both the environment and human operators, requiring safe physical human-robot interaction and being compliant with respect to external forces exerted on its structure (He et al., 2020; Sharifi et al., 2022a).

Motivated by these requirements, we propose in this paper a unifying control architecture suitable for a mobile dual-arm robotic system that allows handling several kinematic constraints, imposing a desired admittance behavior to the end-effectors, and generating internal motions to maximize different objective functions.

Our approach explicitly integrates pHRI mechanisms within the control hierarchy, ensuring that the robot can safely and intuitively interact with human operators when needed. This is achieved through an admittance control scheme embedded in a Hierarchical Quadratic Programming (HQP) framework, where interaction forces measured at the robot's end-effectors directly influence its motion in a compliant manner. Notably, admittance constraints involving the end-effectors' acceleration are effectively included in the HQP formulation, which outputs desired joint velocities as common for many off-the-shelf robots that do not offer a joint torque control interface. The HQP framework also exploits the Control Barrier Function (CBF) theory to impose strict safety constraints, preventing self-collisions and unsafe interactions with the environment.

As mentioned above, the setting addressed in this paper is within precision agriculture. In deatil, motivated by the ongoing European project CANOPIES[1], which aims to develop novel paradigms for human-multi-robot collaboration and interaction in agricultural contexts, the harvesting task in vineyards is proposed as a validation scenario. Due to the inherent complexity of this task, caused by imperfect perception, occlusions, and high dexterity demands, the assistance of human operators is often necessary. A bi-manual mobile robot (see Figure 1), equipped with several sensors and advanced perception and control capabilities, is adopted for this purpose. A key feature of our approach is the ability to adapt the robot's level of autonomy based on real-time perception feedback and human intervention. In fully autonomous mode, the robot detects and harvests grape bunches based on RGB-D sensor data. However, when perception uncertainty is too high due to occlusions or inaccurate peduncle detection, the system changes to a hand-guiding control mode, allowing the human operator to physically guide the robot's arms. In this mode, the admittance controller enables the end-effectors to move in response to applied forces, while the HQP framework still enforces constraints to ensure safe operation. Once the peduncle of the bunch to harvest is correctly aligned with the cutting tool, the control switches back to autonomous mode for the cutting and depositing phases. In this framework, the level of autonomy is embedded in the HQP controller, with tasks in the hierarchy adjusted to allow either the robot or the human to take the lead, depending on the performance of the perception system.

Summarizing, the main contributions of the paper are listed as follows:

- Development of an HQP-based control architecture for dual-arm mobile robots in precision agriculture, integrating equality and inequality constraints for robust task execution. The framework incorporates Control Barrier Functions (CBFs) to enhance safety by preventing self-collisions and enabling controlled human-robot interaction. Additionally, admittance control allows physical human-robot interaction while ensuring task feasibility and compliance with safety constraints;
- Integration of adaptive autonomy levels, allowing the robot to seamlessly transition between fully autonomous and semi-autonomous operation. This feature enables effective collaboration with human operators by dynamically adjusting the robot's behavior based on real-time perception data and task progress;
- Validation in real-world agricultural scenarios, with extensive testing performed in both laboratory settings and actual vineyard environments. The system was demonstrated in autonomous and semi-autonomous harvesting tasks, highlighting its adaptability and effectiveness in real-world applications.

The paper is organized as follows. Section 2 reviews related work. Section 3 introduces the mathematical tools used in this study, while Section 4 describes the problem addressed in this paper. Section 5 details the proposed solution, and Section 6 discusses simulation and experimental results.

## 2. Related Works

When a robotic system has more DOFs than the ones strictly required for the accomplishment of a specific task, as the ones taken into consideration in this work, it is called *kinematically redundant* for that task and this redundancy has to be properly handled in the computation of the control input. In literature this problem is well-known as *redundancy resolution*, and it has been a topic that gathered the efforts of several researchers for decades.

[1] www.canopies-project.eu

The most common approach is to exploit the additional DOFs to make the robotic system capable of performing multiple tasks simultaneously. More in detail, to solve the possible conflicts that can arise among the tasks, it is often recommendable to assign different priority levels to the tasks and compute the joint velocity that fulfills the obtained hierarchy at best. Among the first works in this direction, it is worth mentioning (Slotine and Siciliano, 1991; Nakamura et al., 1987; Khatib, 1987), in which the authors proposed the so-called *null-space projection* technique, i.e. to project the joint velocities coming from the fulfillment of the lower-priority tasks onto the null space of the Jacobian matrices of the higher-priority ones. In (Chiaverini, 1997), the author proposed a similar method that solves the issue of the occurrence of algorithmic singularities, that has been extended for multiple tasks in (Antonelli, 2009). In (Zanchettin and Rocco, 2012) the authors proposed a user-oriented framework in which the selection of the redundancy resolution criterion is decoupled from the Closed-Loop Inverse Kinematics (CLIK) algorithm, giving more flexibility to the control architecture. An overview and comparison of different null-space projection techniques is conducted in (Dietrich et al., 2015). More recently, these kinds of control frameworks have been extended to give the possibility to include in the hierarchy control objectives aimed at keeping the task value within desired thresholds rather than assuming a specific value (Flacco et al., 2015). These tasks are referred to as *inequality constraints* (Simetti and Casalino, 2016) or *set-based tasks* (Di Lillo et al., 2021a). Other more recent and notable works in which the authors spent efforts in searching for a general control framework flexible enough to work with different control interfaces and types of constraints are described in (Fiore et al., 2023; Kazemipour et al., 2022).

A very popular alternative to the null-space projection technique is the Hierarchical Quadratic Programming control framework, in which the joint velocity that fulfills the hierarchy is obtained from the solution of a cascade of Quadratic Programming (QP) problems (Escande et al., 2014). The computational speed of modern QP solvers made this approach suitable for highly redundant systems, such as humanoids (Lee et al., 2021; Kim et al., 2022), aerial vehicle-manipulator systems (Rossi et al., 2017) or even multi-robot systems (Koung et al., 2021), guaranteeing a good scaling of the execution time varying the number of constraints taken into account with respect to other task-priority frameworks. Moreover, the versatility of the approach has led to several works that address the continuity of the solution in case of switching tasks in the hierarchy, as, e.g. in (Kim et al., 2019), weighting of the solution to prefer some DOFs with respect to others (Jang et al., 2022), and handling singularities (Hong et al., 2021). Compared to traditional null-space projection techniques, the HQP formulation has the advantage of intrinsically handling inequality constraints, facilitating the formulation of safety constraints that the system has to respect. In this perspective, a powerful mathematical tool designed to guarantee the safety of a system is the Control Barrier Functions (CBFs) (Ames et al., 2019; Ferraguti et al., 2022), which are often combined with Control Lyapunov Functions (CLFs) (Gehlhar and Ames, 2021) in a single QP problem achieving the so-called *safety-critical control* (Basso and Pettersen, 2020; Nguyen and Sreenath, 2022). Other advantages of the HQP control framework with respect to widely-used task-priority algorithms are the possibility to directly include in the hierarchy constraints involving the decision variables (i.e., joint velocities or joint torques), and to generate always a continuous solution, while traditional task-priority control frameworks have to leverage the design of ad-hoc smooth functions during the activation of high-priority constraints (Simetti and Casalino, 2016; Kim et al., 2019) to avoid discontinuities.

Regarding embedding shared control in robot control architectures in the context of physical human-robot interaction, several approaches have been proposed in the last decades (Losey et al., 2018). For instance, the work in (Losey and O'Malley, 2018) modulates the robot's role in terms of leadership by adjusting the robot's trajectory in response to the exerted human forces. To accomplish this, an energy-based constrained optimization problem is formulated that allows the human to modulate both the actual and desired trajectories of the robot. An alternative solution to a similar issue is outlined in (Balachandran et al., 2020). Specifically, control allocation is established through a metric derived from a Bayesian filter, which dynamically adjusts with real-time sensor inputs, taking into consideration noise patterns in measurements. Additionally, a stability analysis is provided for the suggested shared control framework, accounting for potential communication delays between the human operator and the robot. In (Hagenow et al., 2021), the focus is on tasks that involve interactions with the environment. The authors introduce the concept of *corrective shared autonomy* where users adjust robot state variables such as positions, forces, and execution rates, building upon an autonomous task model. The study demonstrates the feasibility and benefits of the devised approach, highlighting reduced user effort.

The research outlined in (Chen and Ro, 2022) explores variable admittance while accommodating the adaptation to human intentions and meeting system passivity to ensure a safe and intuitive human-robot interaction. The study introduces the notion of *power envelope regulation* to impose limitations on variable admittance parameters deduced from human intentions, thus enhancing the safety of the interaction.

Finally, addressing passivity in mixed-initiative shared control is also explored in (Balachandran et al., 2023). The authors highlight the critical issue that altering forces without maintaining consistent velocity scaling can endanger the system's passivity. Therefore, they redefine the concept of adaptive mixed-initiative shared control as an adaptive stiffness control strategy aimed at preserving passivity while scaling forces.

In this paper, we propose a flexible framework for controlling a redundant dual-arm robotic system using a Hierarchical Quadratic Programming approach. Tasks are organized into three functional categories with descending priority. At the highest priority level are the safety tasks, formulated using CBFs theory and integrated into the HQP framework. The next level encompasses operational tasks, which manage the end-effectors' motion by assigning various admittance behaviors, ensuring compliance with external forces. The lowest priority is given to optimization tasks, which generate internal motions to enhance the robot's posture. It is shown that this comprehensive control framework is well-suited also for pHRI applications and has been extensively tested in experiments with a robot designed for precision agriculture settings. The authors believe that, while the proposed control architecture integrates existing tools, it constitutes a significant contribution due to several factors. These include the complexity of the utilized robotic platform, its versatility, the experimental setup employed to validate the overall design, and its potential applicability beyond precision agriculture into other domains.

## 3. Mathematical Background

In this section, we provide the required mathematical background regarding the robot kinematic modeling, the adopted HQP formulation and the CBF theory that will be used in the following sections to build the proposed control architecture.

### 3.1. Robot Kinematics

Let us consider a mobile robot provided with a movable torso and a dual-arm system (see Figure 1). The state of the system is the vector stacking all the joint positions vector that can be represented as:

$$\boldsymbol{q} = \begin{bmatrix} \boldsymbol{q}_b^T & \boldsymbol{q}_t^T & \boldsymbol{q}_l^T & \boldsymbol{q}_r^T \end{bmatrix}^T \in \mathbb{R}^n, \tag{1}$$

where $\boldsymbol{q}_b \in \mathbb{R}^{n_b}$ is the vector gathering joint variables related to the mobile base, $\boldsymbol{q}_t \in \mathbb{R}^{n_t}$ is the vector gathering joint variables related to the torso, $\boldsymbol{q}_l \in \mathbb{R}^{n_a}$ is the vector of joint variables related to the left arm, $\boldsymbol{q}_r \in \mathbb{R}^{n_a}$ is the vector of the joint variables related to the right arm, and $n = n_b + n_t + 2n_a$ is the number of degrees of freedom (DOFs) of the entire system.

Let us denote with $\boldsymbol{x}_i = \begin{bmatrix} \boldsymbol{p}_i^T & \boldsymbol{o}_i^T \end{bmatrix}^T \in \mathrm{SE}(3)$ the configuration vector of the end-effector $i$, where $\boldsymbol{p}_i \in \mathbb{R}^3$ is the end-effector position vector and $\boldsymbol{o}_i = \begin{bmatrix} o_{i,1} & o_{i,2} & o_{i,3} & o_{i,4} \end{bmatrix}^{\mathrm{T}} = \begin{bmatrix} \kappa_{o_i} & \boldsymbol{\varrho}_{o_i}^{\mathrm{T}} \end{bmatrix}^{\mathrm{T}} \in \mathrm{SO}(3)$ represents the unit quaternion expressing the orientation, being $\kappa_{o_i}$ the scalar component and $\boldsymbol{\varrho}_{o_i} \in \mathbb{R}^3$ the vector component, and $i \in \{l, r\}$ denotes the quantities associated with left and right arm, respectively.

The Cartesian configuration of the dual-arm system can be expressed by the vector stacking the configuration of both the end-effectors, as $\boldsymbol{x} = \begin{bmatrix} \boldsymbol{x}_l^T & \boldsymbol{x}_r^T \end{bmatrix}^T \in \mathrm{SE}(3) \times \mathrm{SE}(3)$. Additionally, by denoting with $\boldsymbol{v}_i = \begin{bmatrix} \dot{\boldsymbol{p}}_i^T & \boldsymbol{\omega}_i^T \end{bmatrix}^T \in \mathbb{R}^6$ the velocity vector of the end-effector $i$, where $\dot{\boldsymbol{p}}_i \in \mathbb{R}^3$ is the end-effector linear velocity vector and $\boldsymbol{\omega}_i \in \mathbb{R}^3$ is the end-effector angular velocity vector, the dual-arm system Cartesian velocity can be expressed as $\boldsymbol{v} = \begin{bmatrix} \boldsymbol{v}_l^T & \boldsymbol{v}_r^T \end{bmatrix}^T \in \mathbb{R}^{12}$.

The existing differential relationship between $\boldsymbol{v}$ and the joint velocity vector $\dot{\boldsymbol{q}} = \begin{bmatrix} \dot{\boldsymbol{q}}_b^T & \dot{\boldsymbol{q}}_t^T & \dot{\boldsymbol{q}}_l^T & \dot{\boldsymbol{q}}_r^T \end{bmatrix}^T$ obtained by derivating the state vector of the system in Eq. (1) can be expressed as follows:

$$\boldsymbol{v} = \begin{bmatrix} \boldsymbol{v}_l \\ \boldsymbol{v}_r \end{bmatrix} = \begin{bmatrix} \boldsymbol{J}_l(\boldsymbol{q}) \\ \boldsymbol{J}_r(\boldsymbol{q}) \end{bmatrix} \dot{\boldsymbol{q}} = \boldsymbol{J}(\boldsymbol{q})\, \dot{\boldsymbol{q}}, \tag{2}$$

where $\boldsymbol{J}(\boldsymbol{q}) \in \mathbb{R}^{12\times n}$ is the Jacobian matrix, that can be partitioned as follows $\boldsymbol{J}_l \in \mathbb{R}^{6\times n}$ e $\boldsymbol{J}_r \in \mathbb{R}^{6\times n}$:

$$\boldsymbol{J}(\boldsymbol{q}) = \begin{bmatrix} \boldsymbol{J}_{b,l}(\boldsymbol{q}_b) & \boldsymbol{J}_{t,l}(\boldsymbol{q}_t) & \boldsymbol{J}_{a,l}(\boldsymbol{q}_l) & \boldsymbol{O}_{6\times n_a} \\ \boldsymbol{J}_{b,r}(\boldsymbol{q}_b) & \boldsymbol{J}_{t,r}(\boldsymbol{q}_t) & \boldsymbol{O}_{6\times n_a} & \boldsymbol{J}_{a,r}(\boldsymbol{q}_r) \end{bmatrix}, \tag{3}$$

where $\boldsymbol{O}_{x\times y}$ is a zero matrix of size $(x \times y)$, to highlight that the base and torso joint velocities influence the velocities of both the end-effectors, while $\dot{\boldsymbol{q}}_l$ and $\dot{\boldsymbol{q}}_r$ only the velocity of left and right end-effector, respectively. Finally, the dependence on $\boldsymbol{q}$ of the related quantities is omitted in the following unless strictly required.

### 3.2. Control Barrier Functions

In this section, we provide the basic theory of Control Barrier Functions, which will be used in the following for formalizing several control objectives aimed at assuring constraints such as the safety of the robot (Ames et al., 2019). Let us consider a general system having the following dynamics:

$$\dot{\xi} = \boldsymbol{f}(t, \xi) + \boldsymbol{g}(\xi)\boldsymbol{u}, \tag{4}$$

where $\boldsymbol{f}$ and $\boldsymbol{g}$ are Lipschitz-continuous vector fields, $\boldsymbol{\xi} \in \mathcal{D} \subset \mathbb{R}^l$ and $\boldsymbol{u} \in \mathcal{U} \subset \mathbb{R}^q$ are state and input of the system, respectively. Let the $k$th generic constraint be expressed in the following general form: $h_k(\boldsymbol{\xi}) \geq 0$, where $h_k(\cdot)$ is a continuous differentiable function in the domain $\mathcal{D}$. According to the CBF framework, let $\mathcal{C}_k \subset \mathcal{D}$ be defined as:

$$\begin{aligned} \mathcal{C}_k &= \{\boldsymbol{\xi} \in \mathbb{R}^l : h_k(\boldsymbol{\xi}) \geq 0\}, \\ \partial\mathcal{C}_k &= \{\boldsymbol{\xi} \in \mathbb{R}^l : h_k(\boldsymbol{\xi}) = 0\}, \\ \mathrm{Int}(\mathcal{C}_k) &= \{\boldsymbol{\xi} \in \mathbb{R}^l : h_k(\boldsymbol{\xi}) > 0\}, \end{aligned} \tag{5}$$

implying that the state $\boldsymbol{\xi}$ is required to belong to the set $\mathcal{C}_k$ in order to satisfy constraint $k$. Function $h_k$ is a CBF if an extended class $\mathcal{K}_\infty$ function $\alpha_k$ exists such that, for a dynamic system represented as in Eq. (4), it holds:

$$\sup_{\boldsymbol{u}\in\mathcal{U}} \left[ L_f h_k(\boldsymbol{\xi}) + L_g h_k(\boldsymbol{\xi})\boldsymbol{u} \right] \geq -\phi_k \alpha_k(h_k(\boldsymbol{\xi})), \tag{6}$$

where $\phi_k > 0$, and $L_f h_k$ and $L_g h_k$ are the Lie derivatives of function $h_k$ with respect to $f$ and $g$, respectively. Then, the following theorem holds (Ames et al., 2019).

**Theorem 1.** *Let function $h_k : \mathcal{D} \subset \mathbb{R}^l \rightarrow \mathbb{R}$ be a continuously differentiable function and the corresponding set $C_k$ defined as in Eq. (5). If $h_k$ is a CBF on $\mathcal{D}$ and $\frac{\partial h_k}{\partial \boldsymbol{\xi}}(\boldsymbol{\xi}) \neq 0$ $\forall \boldsymbol{\xi} \in \partial C_k$, then any Lipschitz continuous controller $u(\boldsymbol{\xi})$ for system in Eq. (4) satisfying Eq. (6) renders the set $C_k$ asymptotically stable.*

*Proof.* The proof can be found in (Ames et al., 2019). □

Remarkably, since Eq. (6) is affine in the control input $\boldsymbol{u}$, the latter can be computed as the result of a convex optimization problem subject to the constraint:

$$\begin{aligned} \boldsymbol{u}^\star &= \arg\min_{\boldsymbol{u}} \frac{1}{2}\left(\boldsymbol{u}-\boldsymbol{u}_{(\cdot)}\right)^{\mathrm{T}} \boldsymbol{Q}\left(\boldsymbol{u}-\boldsymbol{u}_{(\cdot)}\right) \\ \text{s.t.} \quad & \sup_{\boldsymbol{u}\in\mathcal{U}} \left[L_f h_k(\boldsymbol{\xi}) + L_g h_k(\boldsymbol{\xi})\boldsymbol{u}\right] \geq -\phi_k h_k(\boldsymbol{\xi}), \quad \forall\, k \end{aligned} \tag{7}$$

where $\boldsymbol{u}_{(\cdot)}$ is any nominal input for the system and $\boldsymbol{Q} \in \mathbb{R}^{q\times q}$ is a positive definite weight matrix.

### 3.3. Hierarchical Quadratic Programming

The HQP control framework computes the control signal that optimally fulfills a given task hierarchy, minimizing the error of lower-priority tasks when full accomplishment is constrained by higher-priority ones. The HQP formulation converts this hierarchy into a series of Quadratic Programming (QP) problems (Escande et al., 2014). In this setup, the solution to the $i$-th problem (associated with the task of priority $i$) is obtained by treating the solutions of the QP problems for higher-priority tasks as additional constraints. It is worth noticing that in (Escande et al., 2014), the result of this procedure is proven to be equal to the traditional null-space projection techniques in (Slotine and Siciliano, 1991; Nakamura et al., 1987), at least for task hierarchies containing only equality constraints.

More in-depth, let us consider a task $\boldsymbol{\sigma}_1$ that is related to the system velocity vector $\dot{\boldsymbol{q}}$ through the Jacobian matrix $\partial\boldsymbol{\sigma}_1 \backslash \partial\boldsymbol{q} = \boldsymbol{J}_1(\boldsymbol{q})$, and the minimum and maximum desired task velocity $\underline{\boldsymbol{b}}_1$ and $\bar{\boldsymbol{b}}_1$, respectively. The associated QP problem that computes the solution that fulfills the task can be expressed as:

$$\begin{aligned} \min_{\boldsymbol{w}_1,\dot{\boldsymbol{q}}} \quad & \frac{1}{2}\dot{\boldsymbol{q}}^T \boldsymbol{Q}_1 \dot{\boldsymbol{q}} + \boldsymbol{w}_1^T \boldsymbol{Q}_{w_1} \boldsymbol{w}_1 \\ \text{s.t.} \quad & \underline{\boldsymbol{b}}_1 \leq \boldsymbol{J}_1(\boldsymbol{q})\dot{\boldsymbol{q}} + \boldsymbol{w}_1 \leq \bar{\boldsymbol{b}}_1 \end{aligned} \tag{8}$$

where $\boldsymbol{w}_1$ is a *slack variable* vector associated with the task $\boldsymbol{\sigma}_1$ to relax its constraint in case of non-feasibility of the task, $\boldsymbol{Q}_1$ and $\boldsymbol{Q}_{w_1}$, instead, are the cost matrices associated with the decision variables vector $\dot{\boldsymbol{q}}$ and the slack variables vector $\boldsymbol{w}_1$, respectively. It is worth noticing that in case $\underline{\boldsymbol{b}}_1 = \bar{\boldsymbol{b}}_1 = \boldsymbol{b}_1$, the bilateral constraint in Eq. (8) is equivalent to an equality constraint (that is $\boldsymbol{J}_1(\boldsymbol{q})\dot{\boldsymbol{q}} + \boldsymbol{w}_1 = \bar{\boldsymbol{b}}_1$).

Let us now consider the case in which there is also a task $\boldsymbol{\sigma}_2$ (with related Jacobian $\boldsymbol{J}_2(\boldsymbol{q})$) to be performed with a strict lower priority than the task $\boldsymbol{\sigma}_1$. In this case, to obtain the solution is necessary to solve two separate QP problems. The first one will be structured as (8), the second one, instead, as follows:

$$\begin{aligned} \min_{\boldsymbol{w}_2,\dot{\boldsymbol{q}}} \quad & \frac{1}{2}\dot{\boldsymbol{q}}^T \boldsymbol{Q}_2 \dot{\boldsymbol{q}} + \boldsymbol{w}_2^T \boldsymbol{Q}_{w_2} \boldsymbol{w}_2 \\ \text{s.t.} \quad & \underline{\boldsymbol{b}}_1 \leq \boldsymbol{J}_1(\boldsymbol{q})\dot{\boldsymbol{q}} + \boldsymbol{w}_1^* \leq \bar{\boldsymbol{b}}_1 \\ & \underline{\boldsymbol{b}}_2 \leq \boldsymbol{J}_2(\boldsymbol{q})\dot{\boldsymbol{q}} + \boldsymbol{w}_2 \leq \bar{\boldsymbol{b}}_2 \end{aligned} \tag{9}$$

where $\boldsymbol{w}_1^*$ is the solution of the first QP problem, while $\boldsymbol{w}_2$ is the slack variables vector associated with the task $\boldsymbol{\sigma}_2$, allowing a solution to be found even when the two tasks are in conflict. The minimization of $\boldsymbol{w}_2$ in the objective function of Eq.(9) aims at reducing the error on the secondary task $\boldsymbol{\sigma}_2$ given the fulfillment of the constraints required by the primary task $\boldsymbol{\sigma}_1$.

The described approach can be generalized and employed to handle a hierarchy consisting of $k$ arbitrary tasks. By iterating the method described above, it will be necessary to solve a cascade of $k$ QP problems, with the $i$-th problem structured as follows:

$$\begin{aligned} \min_{\boldsymbol{w}_i,\dot{\boldsymbol{q}}} \quad & \frac{1}{2}\dot{\boldsymbol{q}}^T \boldsymbol{Q}_i \dot{\boldsymbol{q}} + \boldsymbol{w}_i^T \boldsymbol{Q}_{w_i} \boldsymbol{w}_i \\ \text{s.t.} \quad & \underline{\boldsymbol{b}}_k \leq \boldsymbol{J}_k(\boldsymbol{q})\dot{\boldsymbol{q}} + \boldsymbol{w}_k^* \leq \bar{\boldsymbol{b}}_k \quad \forall k \in 1,\dots,i-1 \\ & \underline{\boldsymbol{b}}_i \leq \boldsymbol{J}_i(\boldsymbol{q})\dot{\boldsymbol{q}} + \boldsymbol{w}_i \leq \bar{\boldsymbol{b}}_i \end{aligned} \tag{10}$$

## 4. Problem description

In this section, we first describe the specific robot taken into consideration in this work, then we present the task addressed in this paper.

### 4.1. Robot description

The robotic platform considered throughout this paper is shown in Figure 1. The platform is specifically designed and developed for the European-funded project CANOPIES and is made up of a tracked mobile base and a dual-arm manipulator system. The mobile base is an Alitrak DCT-450P[2] equipped with an Nvidia Jetson Orin, which is a compact energy-efficient AI supercomputer belonging to the Nvidia Jetson Oring family, an Intel Next Unit of Computing (NUC), and several navigation sensors, in particular, two Ouster OS1-gen64 3D-Lidars, an Elipse-E Inertial Measurement Unit (IMU), and a Septentrio Real Time Kinematic dual antenna Global Positioning System (GPS-RTK).

The dual-arm manipulator is installed on the top-tail part of the tracked mobile base, which is provided with unicycle kinematics ($n_b = 2$), and it is a bimanual robotic system completely designed and developed by PAL-Robotics[3]. It consists of a 2-DoF torso ($n_t = 2$) and two 7-DoF manipulators ($n_a = 7$). Regarding the torso, it includes a prismatic and a revolute joint that provides it with the capability of lifting and rotating. Moreover, it is featured with a speaker and a microphone aimed at promoting and facilitating human-robot collaboration and interaction. Along with the two manipulators, there is a head attached to the torso. It includes two joints, which provide it with the capability of performing pan and tilt movements, and an Intel RealSense D435 RGB-D sensor. The arms, instead, are both equipped with seven revolute joints and are featured with a Force-Torque (FT) sensor at the wrist and an Intel RealSense D435 RGB-D sensor, which is mounted on a plastic support fixed up to the wrist next to the FT-sensor. Moreover, both manipulators are equipped with an end-effector tool consisting of a cutter and a two-finger gripper, which aims to provide the robot with the capability to achieve harvesting activities.

[2] www.alitrak.com/products/dct-450
[3] https://pal-robotics.com/

As mentioned above, the robotic platform was specifically designed and developed for the project CANOPIES, which is an EU-funded project focused on developing novel methodologies and paradigms for making more efficient and effective collaboration and interaction between human workers and multi-robot teams involved in activities like harvesting and pruning in contexts of precision farming of permanent crops, such as harvesting and pruning of table-grape vineyards.

The entire control software architecture was designed and developed under the ROS Noetic Ninjemys[4] framework, which is a set of software libraries and tools usually used in research contexts to build robot applications. The QP problems detailed in the above sections are solved using the Gurobi Optimizer solver[5], which is a commercial optimization solver for mathematical programming freely available for Academic purposes. The validation of the entire architecture was performed by using a commercial laptop equipped with an Intel Core i7 − 12700H CPU (up to 4, 90GHz), 32GB DDR5 RAM and 2TB SSD.

### 4.2. Task description

The main task addressed in this work is the harvesting of grape bunches in vineyards. However, it is worth noting that the presented architecture is general and suitable for a wide range of tasks, such as pruning or quality inspection, and can also be adopted for settings beyond agriculture.

Regarding harvesting tasks, the process is described in detail here. To effectively harvest a grape bunch, the robot must first identify the bunches in its surroundings and accurately estimate the 3D position of the peduncle to cut. This task, which is beyond the scope of this work, is performed in real-time by the perception software developed by one of the project partners and uses data from RGB-D cameras (Coll-Ribes et al., 2023). Once the peduncle's position is estimated, the harvesting procedure begins. A dedicated software module generates the desired end-effector position and orientation trajectories, connecting a sequence of appropriate waypoints using trapezoidal velocity profiles. This allows the robot to: $i$) reach a pre-grasp position at a predefined distance from the detected peduncle; $ii$) reach the grasp position on the peduncle; $iii$) close the gripper to hold the bunch; $iv$) close the scissors to cut the peduncle; $v$) place the bunch into a designated container.

In this procedure, the accuracy of peduncle position estimation is particularly critical given the varying distances between the cameras and the grape bunches, as well as potential occlusions caused by leaves or other grape bunches. For this reason, it is often necessary to perform the estimation procedure from different points of view. The proposed strategy is to use the RGB-D sensors mounted on the wrists for this purpose. Initially, the perception software utilizes RGB-D data from the head-mounted sensor to provide an initial estimate of the bunches' position. If the peduncles are occluded by leaves or other grape bunches, or if the grapes are positioned too far away for a precise estimation by the head camera, one of the end-effectors is moved close to the detected bunch to reattempt and improve the estimation of the peduncle position using the wrist-mounted cameras (see Figure 1). After this second run of the perception software, the aforementioned harvesting procedure is initiated.

However, due to the inherent complexity of this task, the task execution might not succeed. Indeed, the perception module might be unable to locate the peduncle due to occlusions caused by canes and/or leaves even by using multiple cameras and different point of views. Hence, the support of human operators might be frequently required and, in such cases, we envisage that the system might explicitly ask for human intervention. More in detail, we envisage the possibility of changing the level of autonomy of the system by allowing a human operator to physically change the end-effector configuration for reaching a better point of view or directly toward the target. At the same time, it is advisable that the system still keeps handling autonomously a number of tasks that would be challenging to control for the human operator, such as the control objectives related to the safety of the system or the ones regarding the control of the internal configuration. In this way, the human operator is free to focus only on the operational task, leaving all the other needed control objectives to the system.

## 5. Proposed solution

In this section, we provide an overview of the proposed control architecture that allows the implementation of the behavior mentioned above. With reference to Figure 2, it consists of 4 modules: ($i$) the Trajectory Generation; ($ii$) the High-Level Layer; ($iii$) the Low-Level Layer; and ($iv$) the Perception System.

The Low-Level Layer, detailed in Section 5.1, is represented by the HQP controller described in Section 3.3, which works on a hierarchy composed of a number of tasks belonging to three possible categories: the safety tasks, responsible for guaranteeing the integrity of the system; the operational task, which controls the end-effector motion; the optimization task, which generates internal motions to control the joint configuration of the system. At each time step, it is responsible to compute the joint velocity command for the robot that fulfills all the prioritized tasks at best according to a lexicographic order. The High-Level Layer, detailed in Section 5.2, is implemented as a Finite-State Machine (FSM) algorithm that changes the operational task in the HQP controller in function of the performance of the robot in achieving agricultural tasks, while the safety and the optimization tasks always remain the same. This allows switching the level of autonomy of the robot, leveraging the collaboration of a human operator for the end-effector motion, while the robot autonomously keeps handling the safety tasks and the internal joint configuration. The Trajectory Generation module is responsible for the computation of proper references in position and orientation for the end-effectors of the robot, and it is needed when the system performs autonomous operations. Finally, the Perception system is responsible for the grape recognition procedure and the estimation of the 3D position of the peduncle to cut for performing the harvesting procedure (Coll-Ribes et al., 2023). It makes use of the RGB-D data coming from the sensors placed either in the robot head or on the wrists.

[4] http://wiki.ros.org/noetic
[5] https://www.gurobi.com

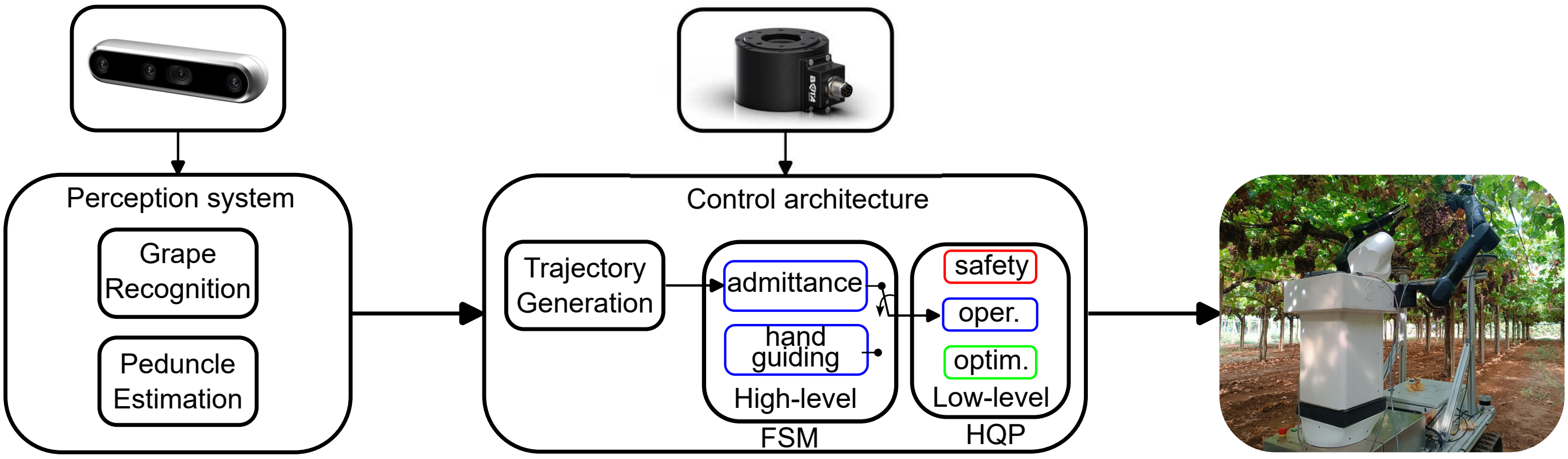

**Figure 2:** Proposed control architecture for the harvesting procedure. The perception system is responsible of the grape detection and of the peduncle position estimation; the Trajectory Generation module generates proper end-effector trajectories for performing the harvesting in an autonomous manner. The High-Level Layer changes the level of autonomy of the robot depending on the performances of the Perception System by changing the operational task in the HQP controller. The Low-Level Layer is responsible for the generation of the joint velocities that fulfill the task hierarchy, and it is implemented as an HQP controller. Finally, these joint velocities go in input to the robot joints internal low-level controllers that generate the actual commands for the motors.

## 5.1. Low-level layer

In this section, all the elementary control objectives that can be combined with different priority levels to achieve various robot behaviors are detailed. Each task is a generic function of the system state $\boldsymbol{\sigma}_x(\boldsymbol{q})$, characterized by a Jacobian matrix $\boldsymbol{J}_x(\boldsymbol{q})$, and might impose an inequality constraint or an equality one depending on the specific control objective to accomplish.

Regarding the tasks that impose an inequality constraint, they are described by minimum and/or admissible maximum values $\underline{\boldsymbol{\sigma}}_x$ and $\bar{\boldsymbol{\sigma}}_x$, they are encoded in proper CBFs $h_x(\boldsymbol{q})$ following the formulation recalled in Section 3.2, and then included as an inequality constraint according to Eq. (10) and the formulation in Section 3.3.

Regarding the tasks that impose equality constraints, it is assumed that a desired trajectory for the task value $\boldsymbol{\sigma}_{x,d}$ is available (and potentially for its velocity $\dot{\boldsymbol{\sigma}}_{x,d}$ and acceleration $\ddot{\boldsymbol{\sigma}}_{x,d}$).

### 5.1.1. *Safety task: Joint position and velocity limits*

In addition to the robot kinematic model reported in Section 3.1, it is worth highlighting that all the joints of the system exhibit physical limits both in position and velocity. These kinematic constraints can be expressed as follows:

$$\begin{cases} \underline{q}_i \leq q_i \leq \bar{q}_i, & i = 1, \cdots, n \\ \underline{\dot{q}}_i \leq \dot{q}_i \leq \bar{\dot{q}}_i, & i = 1, \cdots, n \end{cases} \tag{11}$$

where $\underline{q}_i$ ($\underline{\dot{q}}_i$) represents the minimum position (velocity) that the $i$th joint can achieve, while, in the same way, $\bar{q}_i$ ($\bar{\dot{q}}_i$) represents the maximum position (velocity) value. Since joint limits are constraint by the electromechanical structure of the robot, this task is required to be added at the higher priority in the HQP formulation.

Concerning the joint position limits, it is necessary to exploit the CBF framework to formulate a constraint involving the joint velocity. By defining a task function for the $i$-th joint position as $\boldsymbol{\sigma}_{\mathrm{jp},i} = q_i$ with minimum and maximum admissible values $\underline{\sigma}_{\mathrm{jp},i} = \underline{q}_i$ and $\bar{\sigma}_{\mathrm{jp},i} = \bar{q}_i$, it is possible to express the following functions:

$$\begin{cases} \underline{h}_{\mathrm{jp},i}(q_i) = \sigma_{\mathrm{jp},i} - \underline{\sigma}_{\mathrm{jp},i} \geq 0, & i = 1, \cdots, n \\ \bar{h}_{\mathrm{jp},i}(q_i) = \bar{\sigma}_{\mathrm{jp},i} - \sigma_{\mathrm{jp},i} \geq 0, & i = 1, \cdots, n. \end{cases} \tag{12}$$

Based on Eq. (6), the constraint to enforce for respecting the joint position limits can be expressed as:

$$\begin{cases} \dot{q}_i \geq -\underline{\phi}_{\mathrm{jp},i}\, \underline{h}_{\mathrm{jp},i}(q_i) & i = 1, \cdots, n \\ -\dot{q}_i \geq -\bar{\phi}_{\mathrm{jp},i}\, \bar{h}_{\mathrm{jp},i}(q_i) & i = 1, \cdots, n. \end{cases} \tag{13}$$

with $\bar{\phi}_{\mathrm{jp},i}$, $\underline{\phi}_{\mathrm{jp},i}$ positive scalar gains. Let us define vector functions:

$$\underline{\boldsymbol{h}}_{\mathrm{jp}} = [\underline{h}_{\mathrm{jp},1}, \underline{h}_{\mathrm{jp},2} \cdots, \underline{h}_{\mathrm{jp},n}]^T$$
$$\bar{\boldsymbol{h}}_{\mathrm{jp}} = [\bar{h}_{\mathrm{jp},1}, \bar{h}_{\mathrm{jp},2} \cdots, \bar{h}_{\mathrm{jp},n}]^T$$

and matrices:

$$\underline{\boldsymbol{\Phi}}_{\mathrm{jp}} = \mathrm{diag}\{\underline{\phi}_{\mathrm{jp},1}, \underline{\phi}_{\mathrm{jp},2}, \cdots, \underline{\phi}_{\mathrm{jp},n}\}$$
$$\bar{\boldsymbol{\Phi}}_{\mathrm{jp}} = \mathrm{diag}\{\bar{\phi}_{\mathrm{jp},1}, \bar{\phi}_{\mathrm{jp},2}, \cdots, \bar{\phi}_{\mathrm{jp},n}\}$$

It is straightforward to verify that Eq. (13) can be more conveniently expressed as:

$$\underline{\boldsymbol{b}}_{\mathrm{jp}} = \underline{\boldsymbol{\Phi}}_{\mathrm{jp}}\ \underline{\boldsymbol{h}}_{\mathrm{jp}}(\boldsymbol{q}) \leq \boldsymbol{J}_{\mathrm{jp}}\dot{\boldsymbol{q}} \leq \bar{\boldsymbol{\Phi}}_{\mathrm{jp}}\ \bar{\boldsymbol{h}}_{\mathrm{jp}}(\boldsymbol{q}) = \bar{\boldsymbol{b}}_{\mathrm{jp}}\ , \quad (14)$$

with $\boldsymbol{J}_{\mathrm{jp}} = \boldsymbol{I}_n$, being $\boldsymbol{I}_n$ the $(n \times n)$ identity matrix. This constraint can be easily included in the HQP formulation at the desired priority level as in Eq. (10).

Regarding the joint velocity limits, the constraint to add to the HQP formulation is straightforward. With reference to Eq. (10), defining $\underline{\boldsymbol{b}}_{\mathrm{jv}} = \underline{\dot{\boldsymbol{q}}}$ and $\bar{\boldsymbol{b}}_{\mathrm{jv}} = \bar{\dot{\boldsymbol{q}}}$, the constraint to be added to the $i$-th priority level of the HQP can be written as:

$$\underline{\boldsymbol{b}}_{\mathrm{jv}} \leq \boldsymbol{J}_{\mathrm{jv}}\ \dot{\boldsymbol{q}} \leq \bar{\boldsymbol{b}}_{\mathrm{jv}}\ , \quad (15)$$

where $\boldsymbol{J}_{\mathrm{jv}} = \boldsymbol{I}_n$.

#### 5.1.2. Safety task: Virtual walls

This task allows to keep the distance between a point along the robot structure and a given virtual plane above a certain threshold in order to avoid self-collisions and/or further constraint the workspace of the robot.
Given a point $\boldsymbol{p}_j(\boldsymbol{q})$ along the robot structure and three not aligned points $\boldsymbol{p}^1$, $\boldsymbol{p}^2$ and $\boldsymbol{p}^3$ belonging to the virtual plane, the task function can be expressed as:

$$\sigma_{\mathrm{vw,j}} = \hat{\boldsymbol{n}}^T(\boldsymbol{p}_j - \boldsymbol{p}^1) \quad (16)$$

where $\hat{\boldsymbol{n}}$ is the unit vector normal to the plane that can be obtained as:

$$\hat{\boldsymbol{n}} = \frac{(\boldsymbol{p}^2 - \boldsymbol{p}^1) \times (\boldsymbol{p}^3 - \boldsymbol{p}^1)}{\left\|(\boldsymbol{p}^2 - \boldsymbol{p}^1) \times (\boldsymbol{p}^3 - \boldsymbol{p}^1)\right\|}.$$

The task Jacobian matrix is defined as:

$$\boldsymbol{J}_{\mathrm{vw},j} = -\hat{\boldsymbol{n}}^T \boldsymbol{J}_{p,j}$$

where $\boldsymbol{J}_{p,j}$ is the positional Jacobian relative to the point $\boldsymbol{p}_j$. Defining the minimum distance to be kept between $\boldsymbol{p}_j$ and the virtual wall as $\underline{\sigma}_{\mathrm{vw},j}$, the following CBF can be defined:

$$\underline{h}_{\mathrm{vw},j} = \sigma_{\mathrm{vw},j} - \underline{\sigma}_{\mathrm{vw},j}\ .$$

The constraint needed to enforce the distance to be above the minimum threshold is:

$$\boldsymbol{J}_{\mathrm{vw},j}\ \dot{\boldsymbol{q}} \geq -\underline{\phi}_{\mathrm{vw},j}\ \underline{h}_{\mathrm{vw},j}(\boldsymbol{q})\ , \quad (17)$$

with $\phi_{\mathrm{vw},j}$ a positive scalar gain.

In case there are $L$ points of the serial chain to keep at a minimum distance from the same virtual plane, the constraints can be grouped by stacking the Jacobian matrices $\boldsymbol{J}_{\mathrm{vw},j}$ as:

$$\boldsymbol{J}_{\mathrm{vw}} = \begin{bmatrix}\boldsymbol{J}^T_{\mathrm{vw},1} & \boldsymbol{J}^T_{\mathrm{vw},2} & \cdots & \boldsymbol{J}^T_{\mathrm{vw},L}\end{bmatrix}^T$$

and the CBFs as:

$$\underline{\boldsymbol{h}}_{\mathrm{vw}}(\boldsymbol{q}) = \left[\underline{h}_{\mathrm{vw},1}, \underline{h}_{\mathrm{vw},2}, \cdots, \underline{h}_{\mathrm{vw},L}\right]^T\ .$$

By defining the matrix:

$$\underline{\boldsymbol{\Phi}}_{\mathrm{vw}} = \mathrm{diag}\{\underline{\phi}_{\mathrm{vw},1}, \underline{\phi}_{\mathrm{vw},2}, \cdots, \underline{\phi}_{\mathrm{vw},L}\}\ ,$$

the overall constraint can be expressed as:

$$\boldsymbol{J}_{\mathrm{vw}}\ \dot{\boldsymbol{q}} \geq -\underline{\boldsymbol{\Phi}}_{\mathrm{vw}}\ \underline{\boldsymbol{h}}_{\mathrm{vw}}(\boldsymbol{q}) = \underline{\boldsymbol{b}}_{\mathrm{vw}}\ , \quad (18)$$

and included in the HQP framework as in Eq. (10).

#### 5.1.3. Safety task: Self-collisions avoidance

This task allows to avoid collisions among two generic points of the robot, e.g. between a point belonging to one of the arms and any other point belonging to any other part of the robot. In general, the control objective is to keep the distance between the points $\boldsymbol{p}_j(\boldsymbol{q})$ and $\boldsymbol{p}_l(\boldsymbol{q})$ above a certain threshold. The task function can be expressed as:

$$\sigma_{\mathrm{sc},j,l} = \left\|\boldsymbol{p}_j - \boldsymbol{p}_l\right\| \quad (19)$$

with corresponding Jacobian:

$$\boldsymbol{J}_{\mathrm{sc},j,l} = -\hat{\boldsymbol{n}}^T_{j,l}\ (\boldsymbol{J}_{p,j} - \boldsymbol{J}_{p,l})$$

being:

$$\hat{\boldsymbol{n}}^T_{j,l} = \frac{\boldsymbol{p}_j - \boldsymbol{p}_l}{\left\|\boldsymbol{p}_j - \boldsymbol{p}_l\right\|}$$

the unit vector which connects the point $\boldsymbol{p}_l$ with the point $\boldsymbol{p}_j$, and $\boldsymbol{J}_{p,j}$ and $\boldsymbol{J}_{p,l}$ are the position Jacobian of the points $\boldsymbol{p}_j$ and $\boldsymbol{p}_l$, respectively.

Defining as the minimum desired distance as $\underline{\sigma}_{\mathrm{sc},j,l}$ a CBF can be designed as:

$$\underline{h}_{\mathrm{sc},j,l} = \underline{\sigma}_{\mathrm{sc},j,l} - \sigma_{\mathrm{sc},j,l}\ ,$$

and the corresponding constraint can be written as:

$$\boldsymbol{J}_{\mathrm{sc},j,l}\ \dot{\boldsymbol{q}} \geq -\underline{\phi}_{\mathrm{sc},j,l}\ \underline{h}_{\mathrm{sc},j,l}\ , \quad (20)$$

with $\underline{\phi}_{\mathrm{sc},j,l} > 0$. In case there are $n_c$ constraints related to the same point $\boldsymbol{p}_j$, they can be grouped such as follows

$$\boldsymbol{J}_{\mathrm{sc},j} = \begin{bmatrix} \boldsymbol{J}^T_{\mathrm{sc},j,1} & \boldsymbol{J}^T_{\mathrm{sc},j,2} & \cdots & \boldsymbol{J}^T_{\mathrm{sc},j,n_c} \end{bmatrix}^T$$

and the CBFs:

$$\underline{h}_{\mathrm{sc},j} = \begin{bmatrix} \underline{h}_{\mathrm{sc},j,1} & \underline{h}_{\mathrm{sc},j,2} & \cdots & \underline{h}_{\mathrm{sc},j,n_c} \end{bmatrix}^T ,$$

and by defining the matrix:

$$\underline{\boldsymbol{\Phi}}_{\mathrm{sc,j}} = \mathrm{diag}\{\underline{\phi}_{\mathrm{sc},j,1}, \underline{\phi}_{\mathrm{sc},j,2}, \cdots, \underline{\phi}_{\mathrm{sc},j,n_c}\} .$$

Similarly, if there are $M$ constraints related to the same point $\boldsymbol{p}_j$ of the serial chain, piling up the task Jacobian matrices $\boldsymbol{J}_{\mathrm{sc},j}$ such as

$$\boldsymbol{J}_{\mathrm{sc}} = \begin{bmatrix} \boldsymbol{J}^T_{\mathrm{sc},1} & \boldsymbol{J}^T_{\mathrm{sc},2} & \cdots & \boldsymbol{J}^T_{\mathrm{sc},M} \end{bmatrix}^T$$

the CBFs:

$$\underline{\boldsymbol{h}}_{\mathrm{sc}} = \begin{bmatrix} \underline{\boldsymbol{h}}^T_{\mathrm{sc},1} & \underline{\boldsymbol{h}}^T_{\mathrm{sc},2} & \cdots & \underline{\boldsymbol{h}}^T_{\mathrm{sc},M} \end{bmatrix}^T ,$$

and the gain matrices:

$$\underline{\boldsymbol{\Phi}}_{\mathrm{sc}} = \mathrm{diag}\{\underline{\boldsymbol{\Phi}}_{\mathrm{sc},1}, \underline{\boldsymbol{\Phi}}_{\mathrm{sc},2}, \cdots, \underline{\boldsymbol{\Phi}}_{\mathrm{sc},M}\} ,$$

they can be expressed as a single constraint to be included in the HQP framework in Eq. (10) as:

$$\boldsymbol{J}_{\mathrm{sc}}\ \dot{\boldsymbol{q}} \geq -\underline{\boldsymbol{\Phi}}_{\mathrm{sc}}\ \underline{\boldsymbol{h}}_{\mathrm{sc}}(\boldsymbol{q}) = \underline{\boldsymbol{b}}_{\mathrm{sc}} . \tag{21}$$

### 5.1.4. Operational task: Admittance

This task enables the robot end-effectors to follow a desired trajectory while maintaining a compliant behavior in response to external forces that may occur due to contact with a human operator or the environment.

With reference to Figure 2, it is assumed that the desired trajectory for the end-effectors is generated by the Trajectory Generation module as:

$$\begin{aligned} \boldsymbol{a}_d &= \begin{bmatrix} \ddot{\boldsymbol{p}}^T_{l,d} & \boldsymbol{\alpha}^T_{l,d} & \ddot{\boldsymbol{p}}^T_{r,d} & \boldsymbol{\alpha}^T_{r,d} \end{bmatrix}^T \\ \boldsymbol{v}_d &= \begin{bmatrix} \dot{\boldsymbol{p}}^T_{l,d} & \boldsymbol{\omega}^T_{l,d} & \dot{\boldsymbol{p}}^T_{r,d} & \boldsymbol{\omega}^T_{r,d} \end{bmatrix}^T \\ \boldsymbol{\rho}^d &= \begin{bmatrix} \boldsymbol{p}^T_{l,d} & \boldsymbol{o}^T_{l,d} & \boldsymbol{p}^T_{r,d} & \boldsymbol{o}^T_{r,d} \end{bmatrix}^T , \end{aligned}$$

where $\boldsymbol{p}_{(\cdot),d}, \dot{\boldsymbol{p}}_{(\cdot),d}$ and $\ddot{\boldsymbol{p}}_{(\cdot),d}$ are the desired linear position, velocity and acceleration, respectively, while $\boldsymbol{o}_{(\cdot),d}, \boldsymbol{\omega}_{(\cdot),d}$ and $\boldsymbol{\alpha}_{(\cdot),d}$ are the quaternion, angular velocity and acceleration, respectively, for the *r*ight and *l*eft end-effectors.

Assuming the presence of a wrench sensor mounted on the wrist of the manipulators, it is possible to define the vector:

$$\boldsymbol{h} = \begin{bmatrix} \boldsymbol{h}^T_l & \boldsymbol{h}^T_r \end{bmatrix}^T = \begin{bmatrix} \boldsymbol{f}^T_l & \boldsymbol{\mu}^T_l & \boldsymbol{f}^T_r & \boldsymbol{\mu}^T_r \end{bmatrix}^T \tag{22}$$

as the external measured wrench vector stacking the forces $\boldsymbol{f}_{(\cdot)}$ and moments $\boldsymbol{\mu}_{(\cdot)}$ for both the end-effectors. The objective is to have the system exhibit the following dynamics:

$$\boldsymbol{K}_m\ \tilde{\boldsymbol{a}} + \boldsymbol{K}_d\ \tilde{\boldsymbol{v}} + \boldsymbol{K}_p\ \tilde{\boldsymbol{\rho}} = \boldsymbol{h}, \tag{23}$$

where:

$$\tilde{\boldsymbol{a}} = \boldsymbol{a}_d - \boldsymbol{a}, \quad \tilde{\boldsymbol{v}} = \boldsymbol{v}_d - \boldsymbol{v}, \quad \tilde{\boldsymbol{\rho}} = \begin{bmatrix} \boldsymbol{p}_{l,d} - \boldsymbol{p}_l \\ \tilde{\boldsymbol{\varrho}}_l \\ \boldsymbol{p}_{r,d} - \boldsymbol{p}_r \\ \tilde{\boldsymbol{\varrho}}_r \end{bmatrix} \tag{24}$$

are the acceleration, velocity and configuration errors, with $\tilde{\boldsymbol{\varrho}}_l$ and $\tilde{\boldsymbol{\varrho}}_l$ being the vector parts of the quaternions $\boldsymbol{o}_{l,d} * \boldsymbol{o}_l^{-1}$ and $\boldsymbol{o}_{r,d} * \boldsymbol{o}_r^{-1}$, respectively. Additionally,

$$\begin{aligned} \boldsymbol{K}_m &= \begin{bmatrix} \boldsymbol{K}_{m,l} & \boldsymbol{O}_{6\times6} \\ \boldsymbol{O}_{6\times6} & \boldsymbol{K}_{m,r} \end{bmatrix} \in \mathbb{R}^{12\times12} \\ \boldsymbol{K}_d &= \begin{bmatrix} \boldsymbol{K}_{d,l} & \boldsymbol{O}_{6\times6} \\ \boldsymbol{O}_{6\times6} & \boldsymbol{K}_{d,r} \end{bmatrix} \in \mathbb{R}^{12\times12} \\ \boldsymbol{K}_p &= \begin{bmatrix} \boldsymbol{K}_{p,l} & \boldsymbol{O}_{6\times6} \\ \boldsymbol{O}_{6\times6} & \boldsymbol{K}_{p,r} \end{bmatrix} \in \mathbb{R}^{12\times12} \end{aligned}$$

are the desired virtual mass, damping and stiffness, respectively, that can be either constant or time-varying.

A constraint involving the end-effectors acceleration can not be directly included in Eq. (10), thus it has to be first manipulated to be expressed as a constraint on the end-effector velocity.

In order to solve the issue, the acceleration of the end-effectors is approximated as:

$$\boldsymbol{a}(t) = \frac{\boldsymbol{v}(t) - \boldsymbol{v}(t - T_s)}{T_s} , \tag{25}$$

where $\boldsymbol{v}(t)$ is the current velocity (the time dependency on the current time will be removed in the following) and $\boldsymbol{v}(t - T_s)$ is the velocity at the previous time step, being $T_s$ the sample time given the digital implementation of the control law.

In virtue of the above approximation, Eq. (23) becomes:

$$\boldsymbol{K}_m \boldsymbol{a}_d - \boldsymbol{K}_m \frac{\boldsymbol{v}}{T_s} + \boldsymbol{K}_m \frac{\boldsymbol{v}(t - T_s)}{T_s} + \boldsymbol{K}_d\ \bar{\boldsymbol{v}} + \boldsymbol{K}_p\ \tilde{\boldsymbol{\rho}} = \boldsymbol{h} , \tag{26}$$

which can be rewritten as:

$$\left( -\frac{\boldsymbol{K}_m}{T_s} - \boldsymbol{K}_d \right) \boldsymbol{v} = \boldsymbol{h} - \boldsymbol{\gamma} , \tag{27}$$

where:

$$\boldsymbol{\gamma} = \boldsymbol{K}_m\, \boldsymbol{a}_d - \frac{\boldsymbol{K}_m}{T_s}\, \boldsymbol{v}(t - T_s) - \boldsymbol{K}_d\, \boldsymbol{v}_d - \boldsymbol{K}_p\, \tilde{\boldsymbol{\rho}}\ .$$

By folding Eq. (2) in Eq. (27), one finally obtains:

$$\left(\frac{\boldsymbol{K}_m}{T_s} + \boldsymbol{K}_d\right) \boldsymbol{J}\, \dot{\boldsymbol{q}} = -\boldsymbol{\gamma} - \boldsymbol{h}. \tag{28}$$

This expression can be used as a constraint in the HQP control framework as:

$$\boldsymbol{J}_{\mathrm{adm}}\, \dot{\boldsymbol{q}} = \boldsymbol{b}_{\mathrm{adm}}\ , \tag{29}$$

where $\boldsymbol{J}_{\mathrm{adm}} = \left(\frac{\boldsymbol{K}_m}{T_s} + \boldsymbol{K}_d\right) \boldsymbol{J}(\boldsymbol{q})$ and $\boldsymbol{b}_{\mathrm{adm}} = -\boldsymbol{\gamma} - \boldsymbol{h}$.

#### 5.1.5. Operational task: Hand-guiding

The hand-guiding task is designed to give the human full control of the robot's end-effectors by manually guiding them to the desired location. This behavior can be obtained from the admittance task by setting $\boldsymbol{K}_p = \boldsymbol{O}_{12\times12}$ and $\boldsymbol{a}^d = \boldsymbol{v}^d = \boldsymbol{0}$ in Eq. (23), thus leading to the following form of the constraint in the HQP formulation:

$$\boldsymbol{J}_{\mathrm{hg}}\, \dot{\boldsymbol{q}} = \boldsymbol{b}_{\mathrm{hg}} \tag{30}$$

where:

$$\boldsymbol{J}_{\mathrm{hg}} = \left(\frac{\boldsymbol{K}_m}{T_s} + \boldsymbol{K}_d\right) \boldsymbol{J}(\boldsymbol{q})$$

and

$$\boldsymbol{b}_{\mathrm{hg}} = -\frac{\boldsymbol{K}_m}{T_s}\, \boldsymbol{v}(t - T_s)\ - \boldsymbol{h}.$$

#### 5.1.6. Optimization task: Preferred joint configuration

This task allows to control the value of each joint variable, i.e., to keep it as close as possible to a desired value (e.g., the center of mechanical joint limits). In this case, the vector of the desired task velocity to be included as an equality constraint in the HQP formulation can be generally expressed as follows:

$$\boldsymbol{b}_{\mathrm{jc}} = \dot{\boldsymbol{q}}_d + \boldsymbol{K}_{\mathrm{jc}}\tilde{\boldsymbol{q}} \tag{31}$$

where $\dot{\boldsymbol{q}}_d$ is the desired joint velocity vector, $\boldsymbol{K}_{\mathrm{jc}} \in \mathbb{R}^{n\times n}$ is the task gain matrix, and $\tilde{\boldsymbol{q}} = (\boldsymbol{q}_d - \boldsymbol{q})$ is the configuration error. As regards the constraint imposed by this task within the HQP framework, it can be expressed as follows:

$$\boldsymbol{J}_{\mathrm{jc}}\dot{\boldsymbol{q}} = \boldsymbol{b}_{\mathrm{jc}} \tag{32}$$

with $\boldsymbol{J}_{\mathrm{jc}} = \boldsymbol{I}_n$. It is worth remarking that both in this task and in the previous one, gains $\boldsymbol{K}_{(\cdot)}$ can be time varying to account for human preferences and stability issues (Sharifi et al., 2022b). Although variable admittance control of robotic systems is out of the scope of this work, the presented framework allows to handle this case as well.

### 5.2. High-level layer

The high-level layer is in charge of changing the robot autonomy according to the context and in order to increase the effectiveness of the harvesting operation described in Section 4.2. According to Figure 2, two possible control modes are envisioned: *autonomous* mode and *guided* mode. The switching between two modes is shown in Figure 3, which presents a graphical representation of the Finite State Machine diagram that implements the shared control strategy.

In *autonomous* mode, the robot autonomously performs the needed end-effector motions to perform the entire harvesting procedure in case perception succeeds in finding the peduncle. In *guided* mode, part of the harvesting process is managed by the human operator in case of failures of the robotic system. More in detail, in this case, the human operator guides the end-effector towards the point to cut by physically moving the end-effector. Then, the robot control mode switches back to *autonomous*, and it continues the remaining steps of the harvesting procedure on its own. It is worth noticing that the thresholds related to the safety tasks have been experimentally tuned in order to allow the robot to effectively perform the desired operations in all their phases while guaranteeing the integrity of the system. For example, the virtual wall on the mobile base has been placed low enough to allow the robot to perform the release of the harvested grape while, at the same time, avoiding the occurrence of collisions between the arm and the mobile base.

The robot behaviors related to the two control modes are implemented by switching the operational task in the HQP framework. When in *autonomous* mode, the operational task is the Admittance in Section 5.1.4, and the desired mass, damping and stiffness are set to:

$$\begin{aligned}
\boldsymbol{K}_{m,(\cdot)} &= \mathrm{diag}\{\begin{bmatrix} k_m^p \boldsymbol{I}_3 & k_m^o \boldsymbol{I}_3 \end{bmatrix}\} \\
\boldsymbol{K}_{d,(\cdot)} &= \mathrm{diag}\{\begin{bmatrix} k_d^p \boldsymbol{I}_3 & k_d^o \boldsymbol{I}_3 \end{bmatrix}\} \\
\boldsymbol{K}_{p,(\cdot)} &= \mathrm{diag}\{\begin{bmatrix} k_p^p \boldsymbol{I}_3 & k_p^o \boldsymbol{I}_3 \end{bmatrix}\}\ .
\end{aligned}$$

These parameters have to be tuned to ensure that the robot can follow the desired trajectory while exhibiting a sufficiently compliant behavior with respect to unexpected external forces caused by accidental contact with the environment. When in *guided* mode, the operational task in the HQP controller is switched to Hand-Guiding, as in Section 5.1.5, and the parameters in Eq. (30) are as before, except the desired stiffness which is set to $\boldsymbol{K}_{p,(\cdot)} = \boldsymbol{O}_{6\times6}$, and the desired acceleration, velocity and position are all set to zero. This set of parameters allows the human operator to freely move the end-effector to the actual location of the peduncle to cut.

## 6. Validation Results

In this section the overall architecture is validated both in laboratory and real conditions. Table 1 contains the kinematic constraints of the robot expressed in terms of the maximum and minimum positions and velocities.

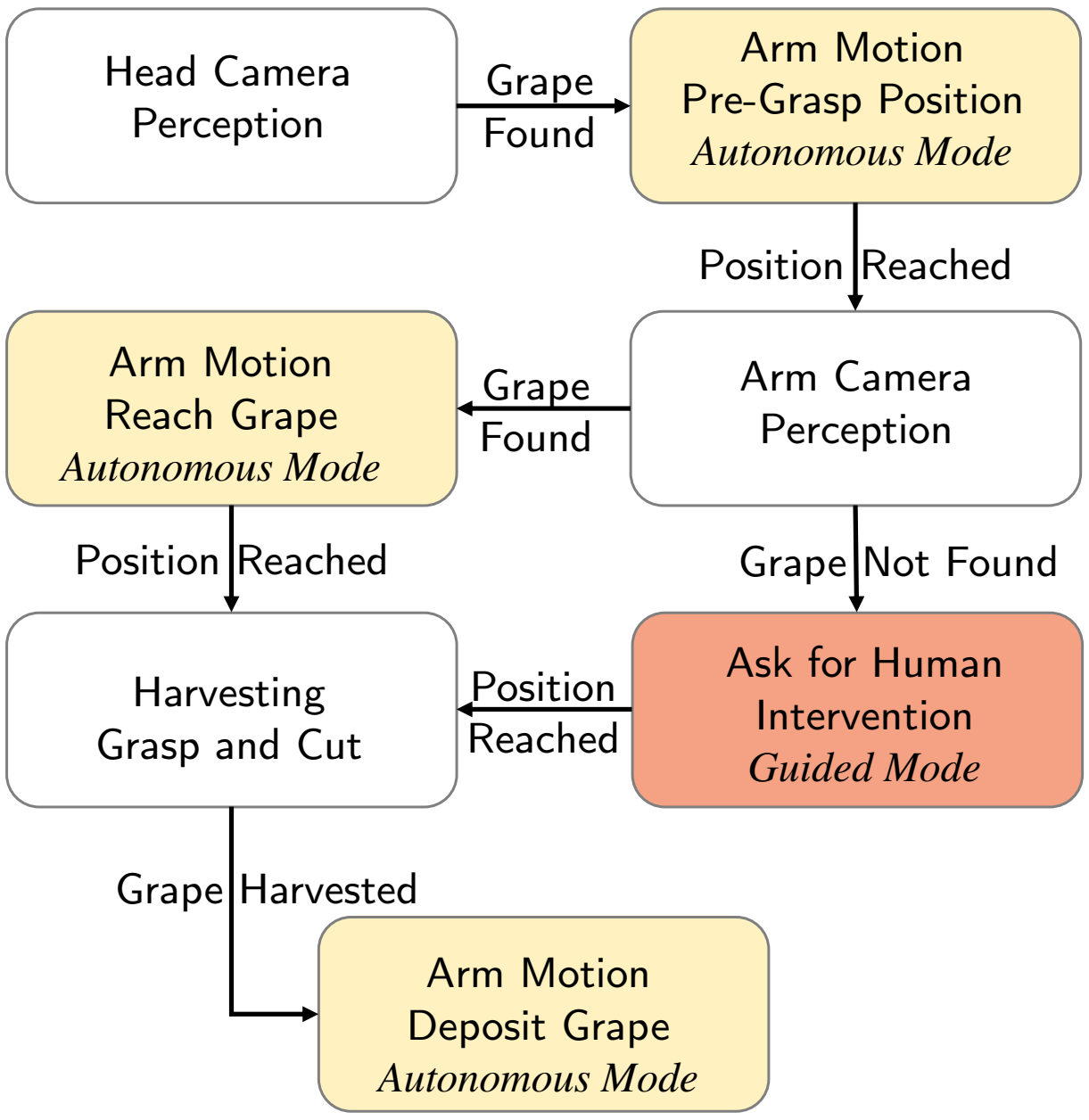

**Figure 3:** State Machine describing the sequence of operations for human-robot collaboration in grape harvesting. The colors of the blocks encode the robot control mode. Yellow: *autonomous*, orange: *guided*. The blocks with white background are related to robot components different from the manipulators (i.e. RGB-D cameras and end-effectors), thus they are not affected by the specific control mode.

| JOINT LIMITS |
|---|
| $\bar{\boldsymbol{q}}_B = [\infty,\ \infty,\ 3.14]^{\mathrm{T}}$<br>$\underline{\boldsymbol{q}}_B = [-\infty,\ -\infty,\ -3.14]^{\mathrm{T}}$ |
| $\bar{\boldsymbol{q}}_T = [3.00,\ 0.30]^{\mathrm{T}}$<br>$\underline{\boldsymbol{q}}_T = [-1.60,\ 0.0]^{\mathrm{T}}$ |
| $\bar{\boldsymbol{q}}_L = [1.44,\ 1.47,\ 2.41,\ 1.49,\ 2.42,\ 2.03,\ 1.28]^{\mathrm{T}}$<br>$\underline{\boldsymbol{q}}_L = [-0.73,\ -1.45,\ -2.41,\ -2.30,\ -2.42,\ -2.03,\ -3.48]^{\mathrm{T}}$ |
| $\bar{\boldsymbol{q}}_R = [1.44,\ 1.47,\ 2.41,\ 1.49,\ 2.42,\ 2.03,\ 1.28]^{\mathrm{T}}$<br>$\underline{\boldsymbol{q}}_R = [-0.73,\ -1.45,\ -2.41,\ -2.30,\ -2.42,\ -2.03,\ -3.48]^{\mathrm{T}}$ |
| $\bar{\dot{\boldsymbol{q}}}_T = [0.7,\ 0.2]^{\mathrm{T}}$<br>$\underline{\dot{\boldsymbol{q}}}_T = [-0.7,\ -0.2]^{\mathrm{T}}$ |
| $\bar{\dot{\boldsymbol{q}}}_L = [1.95,\ 1.95,\ 1.95,\ 1.95,\ 1.95,\ 1.95,\ 1.95]^{\mathrm{T}}$<br>$\underline{\dot{\boldsymbol{q}}}_L = [-1.95,\ -1.95,\ -1.95,\ -1.95,\ -1.95,\ -1.95,\ -1.95]^{\mathrm{T}}$ |
| $\bar{\dot{\boldsymbol{q}}}_R = [1.95,\ 1.95,\ 1.95,\ 1.95,\ 1.95,\ 1.95,\ 1.95]^{\mathrm{T}}$<br>$\underline{\dot{\boldsymbol{q}}}_R = [-1.95,\ -1.95,\ -1.95,\ -1.95,\ -1.95,\ -1.95,\ -1.95]^{\mathrm{T}}$ |

**Table 1**
CANOPIES robot joint position limits (in [m] and [rad]) and joint velocity limits (in [m/s] and [rad/s]).

| PARAMETER | VALUE | EQ. |
|---|---|---|
| $\underline{\phi}_{\mathrm{jp},i}, \bar{\phi}_{\mathrm{jp},i}$ | 10 | Eq.(16)-(17) |
| $\underline{\phi}_{\mathrm{vw},j}$ | 5 | Eq.(25) |
| $\underline{\sigma}_{\mathrm{vw},j,i}$ | $0.3m$ | Eq.(21) |
| $\underline{\phi}_{\mathrm{sc},j,i}$ | 10 | Eq.(32) |
| $\underline{\sigma}_{\mathrm{sc},j,head}$ | $0.5m$ | Eq.(28) |
| $\underline{\sigma}_{\mathrm{sc},j,torso}$ | $0.35m$ | Eq.(28) |
| $\underline{\sigma}_{\mathrm{sc},j,arm}$ | $0.2m$ | Eq.(28) |
| $\boldsymbol{K}_{m,l}, \boldsymbol{K}_{m,r}$ | $\mathrm{diag}\{20\boldsymbol{I}_3,\ 3\boldsymbol{I}_3\}$ | Eq.(43) |
| $\boldsymbol{K}_{d,l}, \boldsymbol{K}_{d,r}$ | $\mathrm{diag}\{253\boldsymbol{I}_3,\ 27\boldsymbol{I}_3\}$ | Eq.(44) |
| $\boldsymbol{K}_{p,l}, \boldsymbol{K}_{p,r}$ | $\mathrm{diag}\{800\boldsymbol{I}_3,\ 60\boldsymbol{I}_3\}$ | Eq.(45) |
| $\boldsymbol{K}_{\mathrm{jc}}$ | $\mathrm{diag}\{2, 0.5, 10\boldsymbol{I}_{14}\}$ | Eq.(54) |

**Table 2**
Value of the parameters used for the validation.

Table 2, instead, reports adopted values for parameters and gains.

## 6.1. Low-level layer validation in laboratory condition

In this section, the results obtained from extensively validating the Low-Level Layer proposed in Section 5.1 in a laboratory environment are reported. A video of the experiments is available at the following link[6].

### *6.1.1. Safety tasks validation*

In the following, we display the results of an experiment performed to validate the safety tasks. The experiment has been carried out by using a hierarchy with three priority levels and five tasks, in detail, four safety tasks and an operational task. The five tasks considered for the experiment have been organized as shown in Figure 4.

Since the aim of this experiment is to validate the safety tasks, we structured a lower-level (third level in the hierarchy) task that generates predefined sinusoidal velocities at each joint of both the arms, intentionally chosen to violate the constraints imposed by the higher-priority safety tasks. To generate these velocities, we exploit the optimization task described in Section 5.1.6, with $\boldsymbol{q}_d$ and $\dot{\boldsymbol{q}}_d$ in Eq. (31) such as:

$$\begin{cases} \boldsymbol{q}_d &= 3\sin\left(\frac{2\pi}{25}t\right)\left[\boldsymbol{0}_{1x3},\quad \boldsymbol{0}_{1x2},\quad 1\quad \dots\quad 1\right]^T \\ \dot{\boldsymbol{q}}_d &= 0.75\sin\left(\frac{2\pi}{25}t\right)\left[\boldsymbol{0}_{1x3},\quad \boldsymbol{0}_{1x2},\quad 1\quad \dots\quad 1\right]^T \end{cases}$$

As anticipated above, the joint position limits have been chosen to mirror the actual limits of the actuators, as reported in Table 1, while the maximum and minimum joint velocities have been intentionally limited at $\bar{\dot{\boldsymbol{q}}} = 0.5$ rad/s and $\underline{\dot{\boldsymbol{q}}} = -0.5$ rad/s to activate the corresponding constraint.

The virtual wall task described in Section 5.1.2 has been used to avoid collisions between the manipulators and the mobile base. Figure 5 contains a representation of the virtual wall with a top-down and a side view of the robot. More in detail, the blue surface represents the virtual wall itself, while the blue spheres represent the points of the manipulator structure subject to the constraint of keeping the distance from the virtual wall greater than the given threshold. For this particular experiment, the minimum distance to keep has been set at $\underline{\sigma}_{\mathrm{vw},j} = 0.3$m.

[6] https://youtu.be/RbsWzz8vnMo

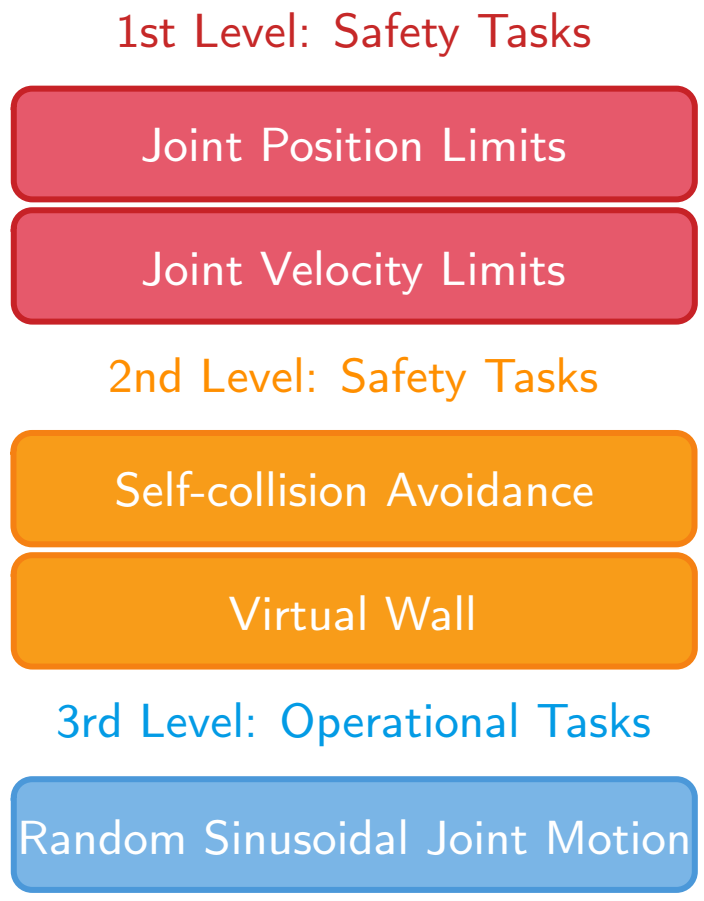

**Figure 4:** The task hierarchy that was used to validate the safety tasks introduced in Sections 5.1.1 - 5.1.3.

The self-collision avoidance task has been used to avoid the occurrence of collisions between the manipulators themselves and between the manipulators and the other components of the bimanual robotic system, i.e. the torso and the head. Specifically, three different thresholds have been defined: for the manipulators $\underline{\sigma}_{\mathrm{sc},j,arm} = 0.2$m, for the head $\underline{\sigma}_{\mathrm{sc},j,head} = 0.5$m, and for the torso $\underline{\sigma}_{\mathrm{sc},j,torso} = 0.35$m. Figure 6 shows representative points interested by the self-collision avoidance tasks. In detail, the images show the robot from different points of view, with interested points encapsulated within spheres and cylinders. In particular, the spheres highlighted in blue indicate the components whose position is conditioned by the constraints related to the task. The spheres and cylinders highlighted in red instead, in addition to indicating the components from which the ones encapsulated within the blue spheres have to be kept at a distance, denoting the regions in which the thresholds associated with the task are not observed.

It is worth noticing that the control of the distance between a point and a cylinder is achieved by exploiting the formulation reported in Section 5.1.3 for the self-collision tasks, but considering as point $\boldsymbol{p}_j$ the closest point between $\boldsymbol{p}_l$ and the cylinder representing the constraint. Table 2 reports the values of all the tasks parameters. In summary, all implemented safety tasks introduce 86 concurrent constraints within the HQP framework. Despite the large number of constraints, the control framework can be effectively executed at 100 Hz, which is the maximum control frequency achievable by the robot actuators.

Figure 7 shows the results obtained from this experiment. In detail, Figures 7.a) - 7.c) show the time evolution of the distances related to the tasks of self-collision avoidance and virtual walls (solid blue lines) with respect to the corresponding threshold values (dashed red lines). Figure 7.d), instead, in the top part reports the time evolution of the normalized joint positions (solid lines) with respect to the corresponding normalized minimum and maximum threshold values (dashed red lines) and in the bottom part, instead, reports the time evolution of joint velocities (solid lines) with respect to the corresponding minimum and maximum threshold values (dashed red lines). By observing these figures, it is possible to notice that throughout the entire experiment, both the joint positions and velocity limits are successfully respected, and no one of the distances related to the tasks of self-collision avoidance and virtual walls takes values below the corresponding threshold.

#### *6.1.2. Operational tasks validation*

In the following, we go through the results of the experiments performed for validating the operational tasks described in Section 5.1.5. In particular, we designed an experiment to validate the concurrent activation of the safety tasks and the hand-guiding operational task. The hierarchy taken into consideration is thus the same as the one described in Section 6.1.1, but this time changing the lower-level task from a sinusoidal joint motion to a hand-guiding task as in Section 5.1.5. The overall hierarchy taken into consideration for this experiment is shown in Figure 8.

In this experiment, a human operator is first asked to manually move the right end-effector in free space, to show the effectiveness of the hand-guiding task; then, the same end-effector is moved in sequence toward the left end-effector, the torso/head and, finally, toward the mobile base, in order to prove the activation of the safety tasks even in this case. Figure 9 displays the results obtained from this experiment. In detail, Figure 9.a) shows the time evolution of the distances related to the right arm for the tasks of self-collision avoidance and virtual walls (solid blue lines) with respect to the corresponding threshold values (dashed red line). Figure 9.b), instead, shows the time evolution of the interaction force applied on the right wrist (top) and of the right end-effector linear velocity (bottom).

It is worth noticing that the raw wrench sensor measurements are pre-processed before being used in Eq. (22). First of all, the end-effector payload is identified using standard least-squares parametric identification to address its influence on the measured wrench (Di Lillo et al., 2020). This process allows compensating for the gripper's load and isolating the actual external wrench. Then, the data-driven thresholding strategy in (Lippi et al., 2021) is employed to distinguish between *contact* and *no contact* situations, which helps prevent the system from reacting to measurement noise that could otherwise unnecessarily disrupt the trajectory tracking.

At $t \approx 10$s the operator starts interacting with the right end-effector that moves accordingly, being the linear velocity different than zero. At $t \approx 20$s the operator moves the right end-effector close to the left end-effectors, triggering the activation of the corresponding self-collision task. For about 5s, the operator pushes it toward the left end-effector (end-effector force is different than zero), but it does not move (end-effector linear velocity equal to zero during). The same happens at $t \approx 33$s and at $t \approx 48s$, when the operator tries to push the end-effector toward the torso and the mobile base. This validates the effectiveness of the implemented task hierarchy, allowing the operator to freely move the end-effector while the robot autonomously handles all the safety tasks.

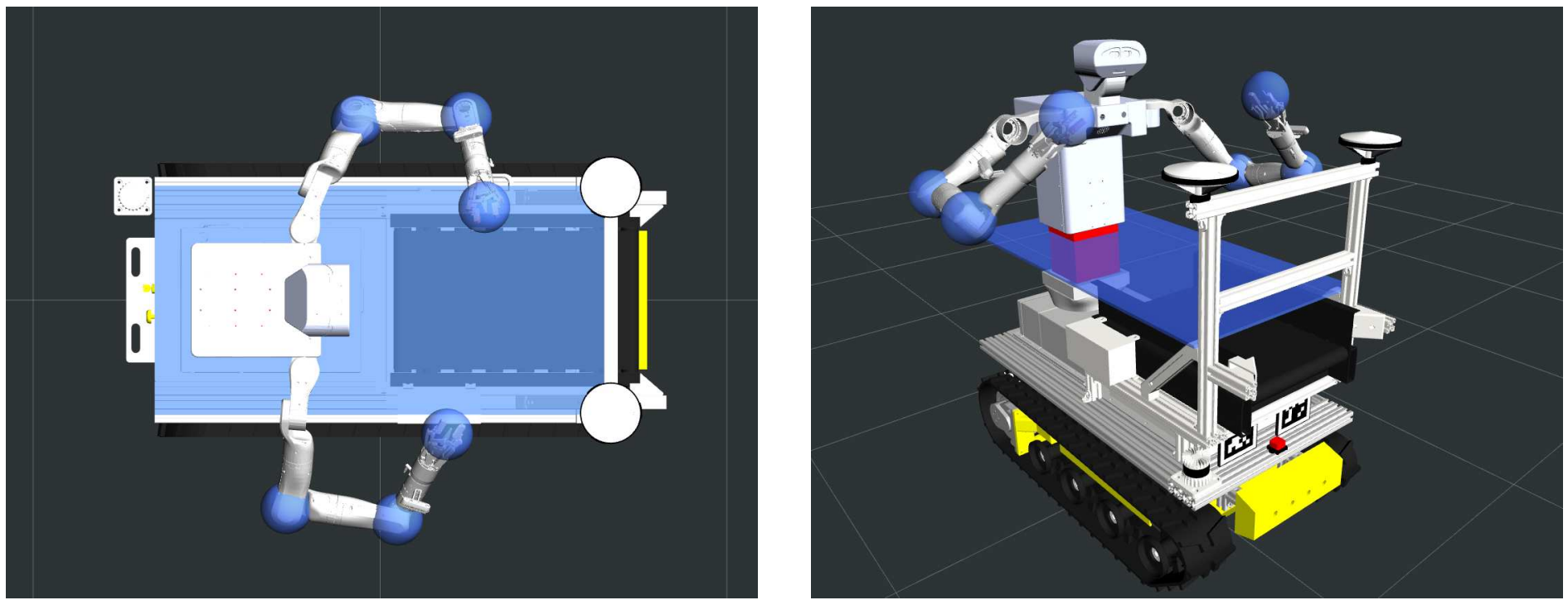

**Figure 5:** Top and side view of the virtual wall task. The horizontal plane over the mobile base is the wall itself, while the points of the manipulators to keep at a minimum distance from the wall are depicted as blue spheres.

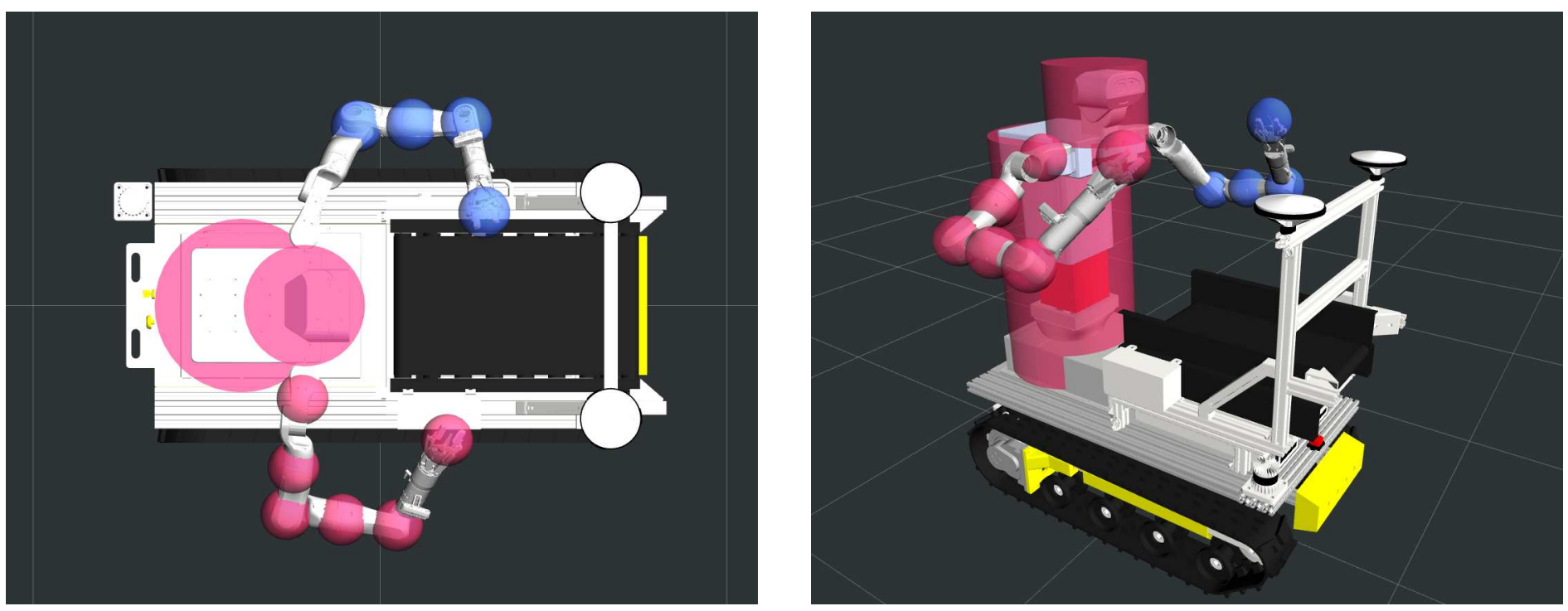

**Figure 6:** Top and side view of the self-collision tasks. All the blue and red spheres along the arms kinematic structure represent points to keep a minimum distance from. The same happens for the boxes surrounding the torso and the head of the robot.

### *6.1.3. Optimization tasks validation*

In the following section, results obtained from the experiments performed to validate the optimization task described in Section 5.1.6 are described. More in detail, two experiments with different task hierarchies were performed. In the first one, the hierarchy described in Section 6.1.2 is taken into account, changing the operational task from hand-guiding to admittance, hence obtaining the hierarchy shown in Figure 10.

In the second one, we consider an additional priority level by introducing the preferred joint posture as optimization task, in order to highlight the difference in the robot behavior. The hierarchy is the one showed in Figure 11.

For both experiments, the desired end-effector position is kept equal to the initial one. The human operator is asked to manually move the end-effector twice and then release it, making it return to the initial position.

Figures 12 and 13 display the results obtained from the first and second experiment, respectively. In detail, Figures 12.a) and 13.a) show the time evolution of the wrenches detected by the F/T sensor installed within the right wrist. Figures 12.b) and 13.b) display the time evolution of the right end-effector pose. Figures 12.c) and 13.c), instead, show the time evolution of the normalized joint positions (solid lines) with respect to the corresponding normalized minimum and maximum threshold values (dashed red lines).

By comparing Figures 12.a and 13.a, it is evident that the interaction forces are very similar in both experiments. Additionally, Figures 12.b and 13.b demonstrate that the behavior of the end-effector is also quite similar, as it returns to the same pose after being released by the operator in both cases. The difference between the two experiments lies in the internal joint configurations in the two cases, as it can be seen in Figures 12.c) and 13.c). More in detail, without including the optimization task in the hierarchy, each time the human operator releases the end-effector the robot reaches the desired pose with a different internal joint configuration. Given the well-known drifting of the inverse kinematics solution in case of redundant manipulators (Klein and Kee, 1989), this behavior can lead to very unnatural and close to the mechanical limits configurations. Including a preferred joint posture as optimization task in the hierarchy helps solving this issue,

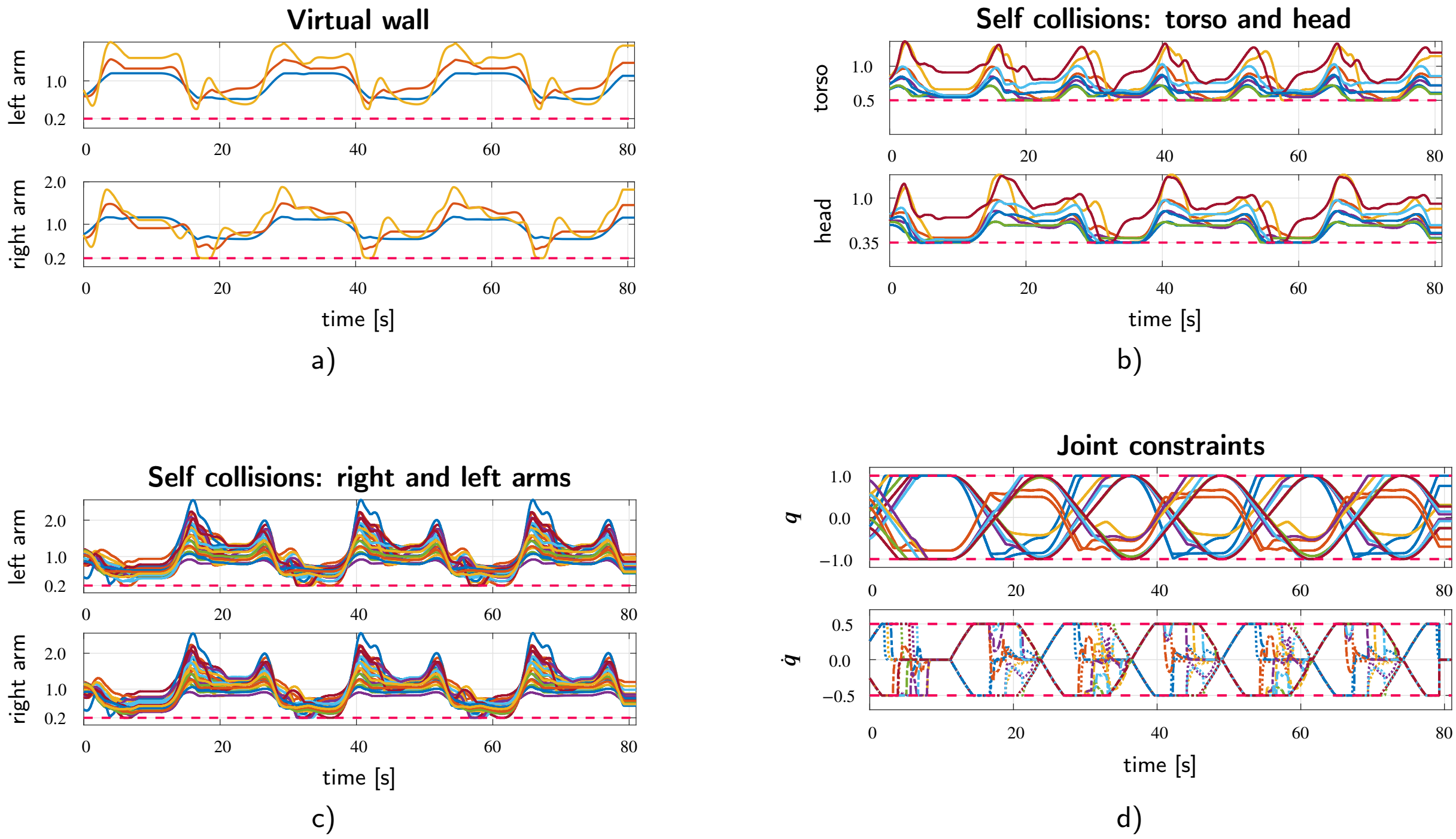

**Figure 7:** Safety tasks validation: a) Distance of three points belonging to the arms from the virtual wall placed over the mobile base and imposed minimum threshold (red-dashed line). b) Distance of three points belonging to the arms from the torso and the head of the robot and imposed minimum threshold (red-dashed line). c) Distance between three points of one of the arm with respect to three points of the other arm and imposed minimum threshold (red-dashed line). d) Normalized joint positions for left (dotted lines) and right (dot-dashed lines) arms and velocity and normalized imposed minimum and maximum thresholds (red-dashed lines).

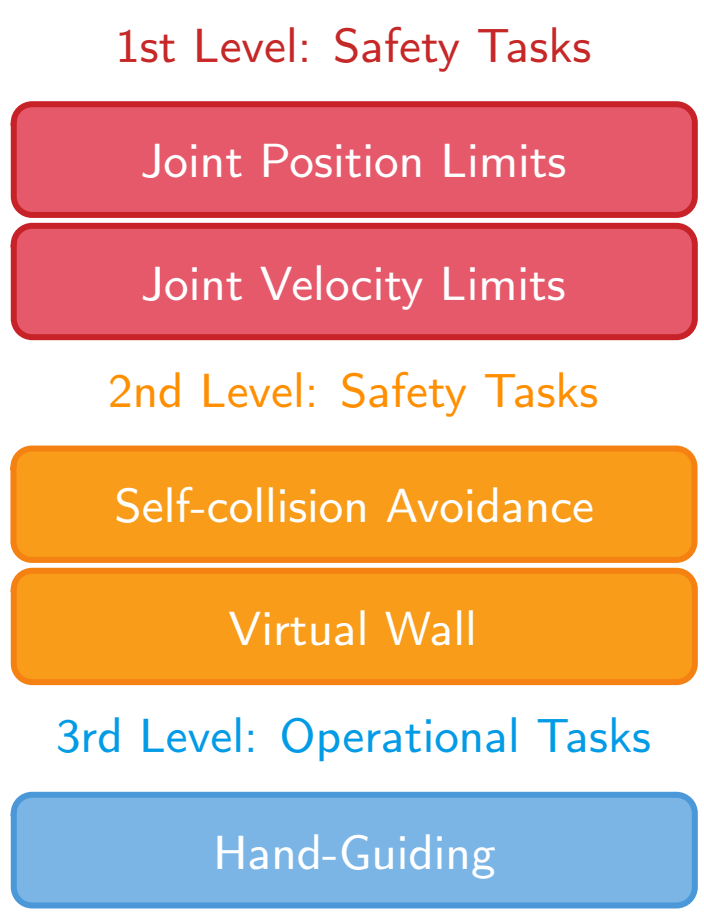

**Figure 8:** The task hierarchy used to validate the operational task described in Section 5.1.5.

as the robot reaches the end-effector poses always with the same internal configurations.

## 6.2. Full architecture validation in real field

In the following sub-sections, the results of the harvesting experiments conducted during the experimental campaign of the CANOPIES project, which was held in Aprilia (Italy) in September 2023 are reported. A video of the experiments in the real filed is available at the following link [7].

### *6.2.1. Autonomous harvesting*

Regarding the harvesting procedure conducted in autonomous mode (see Figure 3), we selected a scenario in which there were no occlusions and the perception system successfully provided a reliable peduncle position estimation.

According to the same FSM in Figure 3, at the beginning of the procedure, the robot is in autonomous mode and runs the perception software to analyze the data collected by the RGB-D sensor placed in its head (Figure 14). The perception system provides the results of the detection and localization process to the control architecture, which then guides the robot to bring its right harvesting tool to the pre-grasping position (Figure 15. Top Right), i.e., at a predefined distance from one of the bunches whose peduncle has successfully been recognized. Then, the robot brings the tool to the grasping position and completes the harvesting procedure, performing the grasp and the cut of the peduncle. Once the grape has been harvested, the robot brings the grape first to the pre-release position (Figure 15. Bottom Left), i.e., at a predefined distance from the box where the grape must be placed, and then to the release position (Figure 15. Bottom Right). Once the grape has been released, the robot brings the tool back to the initial configuration (Figure 15. Top

[7] https://youtu.be/V2dF5SXiKKE

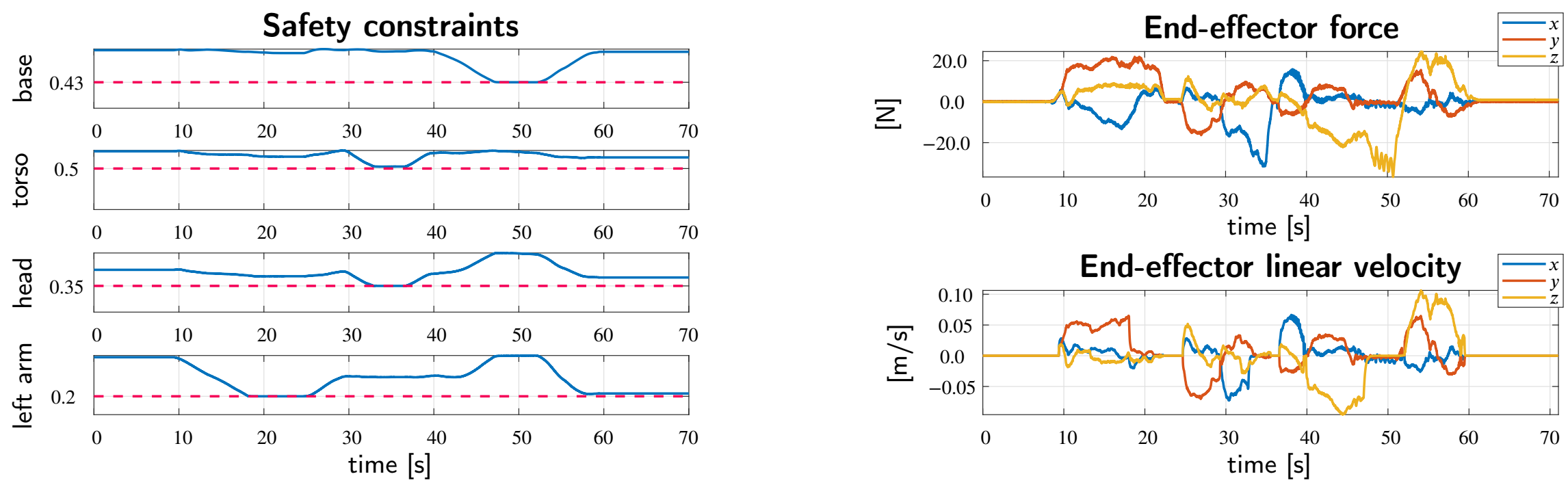

a) b)

**Figure 9:** Hand-guiding operational task validation. a) Distance of the right end-effector from the mobile base, the torso, the head and the left end-effector and imposed minimum distances (red-dashed lines). b) Top: end-effector measured linear force. Bottom: end-effector linear velocity.

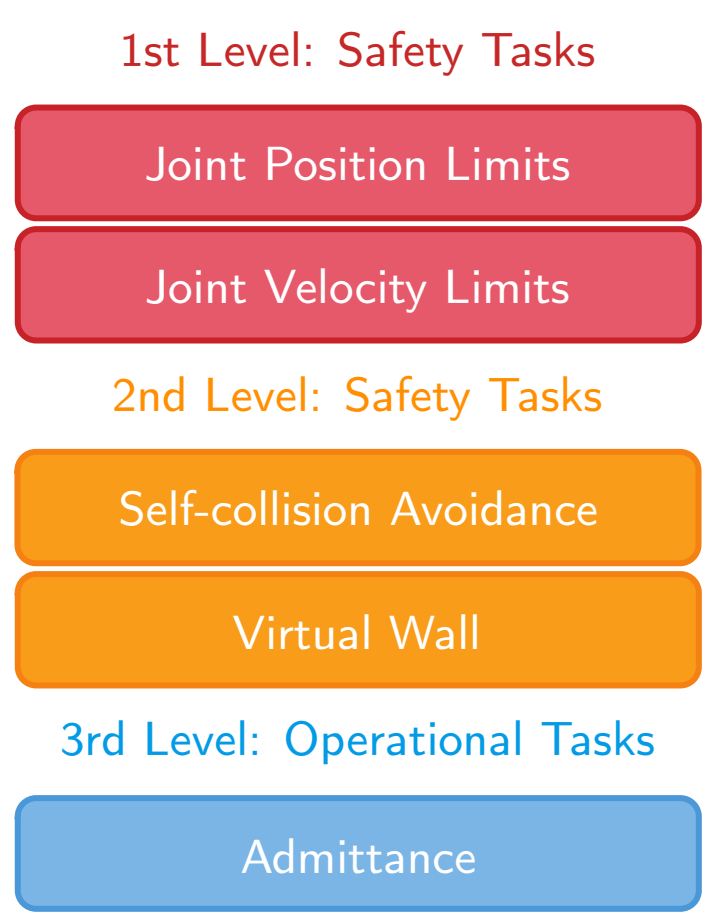

**Figure 10:** First of the two hierarchies that were used to validate the optimization task described in Section 5.1.6.

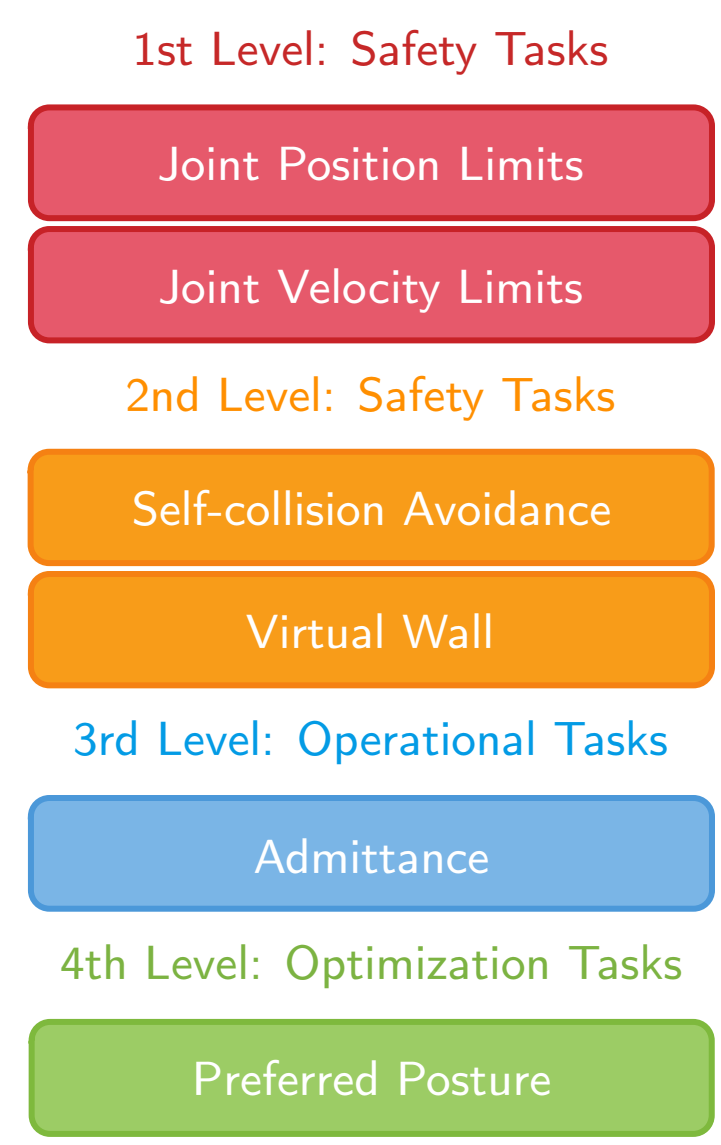

**Figure 11:** Second of the two hierarchies that were used to validate the optimization task described in Section 5.1.6.

Left). Figure 16 (bottom) shows the time evolution of the actual (solid blue line) and the desired (dashed red line) position of the tool involved in the harvesting procedure. Figure 16 (top), instead, shows the time evolution of the normalized positions and velocities of the torso and left arm joints (solid lines) with respect to their normalized limit values (dashed red lines).

### *6.2.2. Semi-autonomous harvesting*

In the following, we report the results of the experiment performed to validate and test the semi-autonmous harvesting procedure described in Section 5.2. In order to validate this control strategy, we selected a scenario in which the robot required operation is to harvest a bunch which is occluded by leaves.

According to the diagram in Figure 3, at the beginning the robot is in *autonomous* control mode and runs the perception software on the data coming from the RGB-D sensor of the head camera (Figure 17. Top Left). Some of the present bunches are detected and localized by the perception system (Figure 17. Top Right); however, peduncles are not clearly visible. Then, the robot moves the right end-effector in a pre-grasping position at a predefined distance from one of the bunch, and makes a new estimation attempt by using the wrist camera (Figure 17. Bottom Left). Even in this case, the peduncle of the target bunch was not recognized and, at this point, the robot asks for the human operator's assistance (by exploiting its text-to-speech module), and its operational task switches to *hand-guiding*, obtaining the same hierarchy reported in Figure 8. The admittance parameters are changed according to this mode, allowing the human operator to grab the end-effector and manually place it on the point to cut.
Figure 18 (top) shows the interaction wrench of the right end-effector during the harvesting experiment. It is possible to notice that from $t \approx 40$s to $t \approx 60$s the human operator exerts a force to move the end-effector. When he/she releases the end-effector, the robot detects a *no contact* situation for more than 3s and it switches back to *autonomous* mode. From that point on, the robot autonomously continues the harvesting procedures by changing the admittance parameters corresponding to the *autonomous* mode. In particular, the two fingers gripper first grasps the peduncle, while the cutter cuts it (approximately at $t \approx 70$s); then, the trajectory generation module plans a trajectory to release the grape into the collecting box placed in the front part of the robot (at $t \approx 130$s). It is worth noticing that, during this phase, the measured interaction wrench is mainly represented by the gravity due to the grape weight (yellow line in the top plot in Figure 18). With regards to joint constraints, Figure 18 (bottom) shows that normalized joint trajectories and velocities for the torso and the right arm (the left arm is here omitted since it is not adopted in the considered experiment) are within their (normalized) limits.

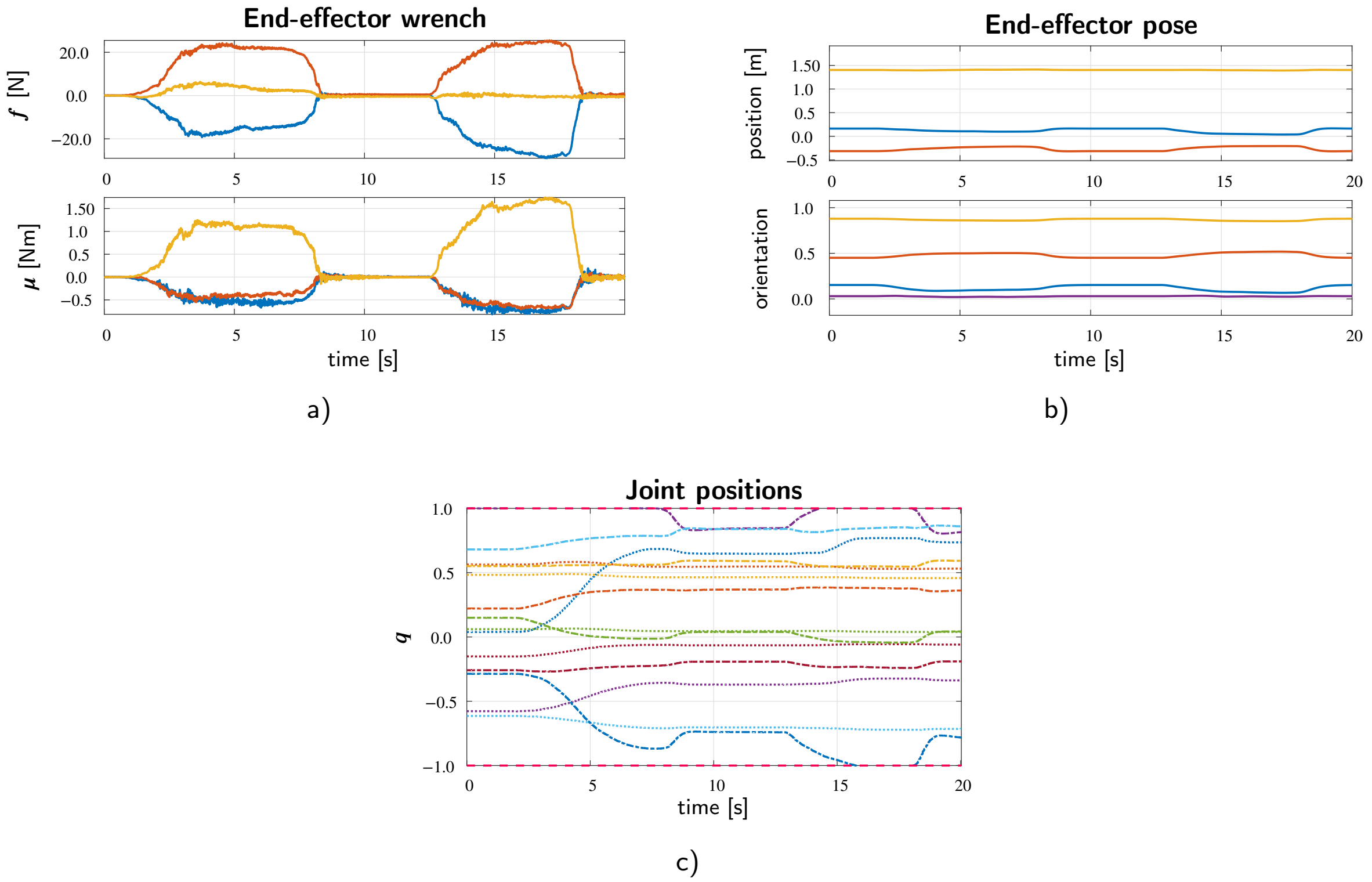

**Figure 12:** Optimization tasks validation without including the optimization task in the hierarchy. a) Top: measured end-effector linear force. Bottom: measured end-effector moment. b) Top: end-effector position. Bottom: end-effector orientation expressed as quaternion. c) Normalized joint positions (dotted line for left arm, dot-dashed line for right arm).

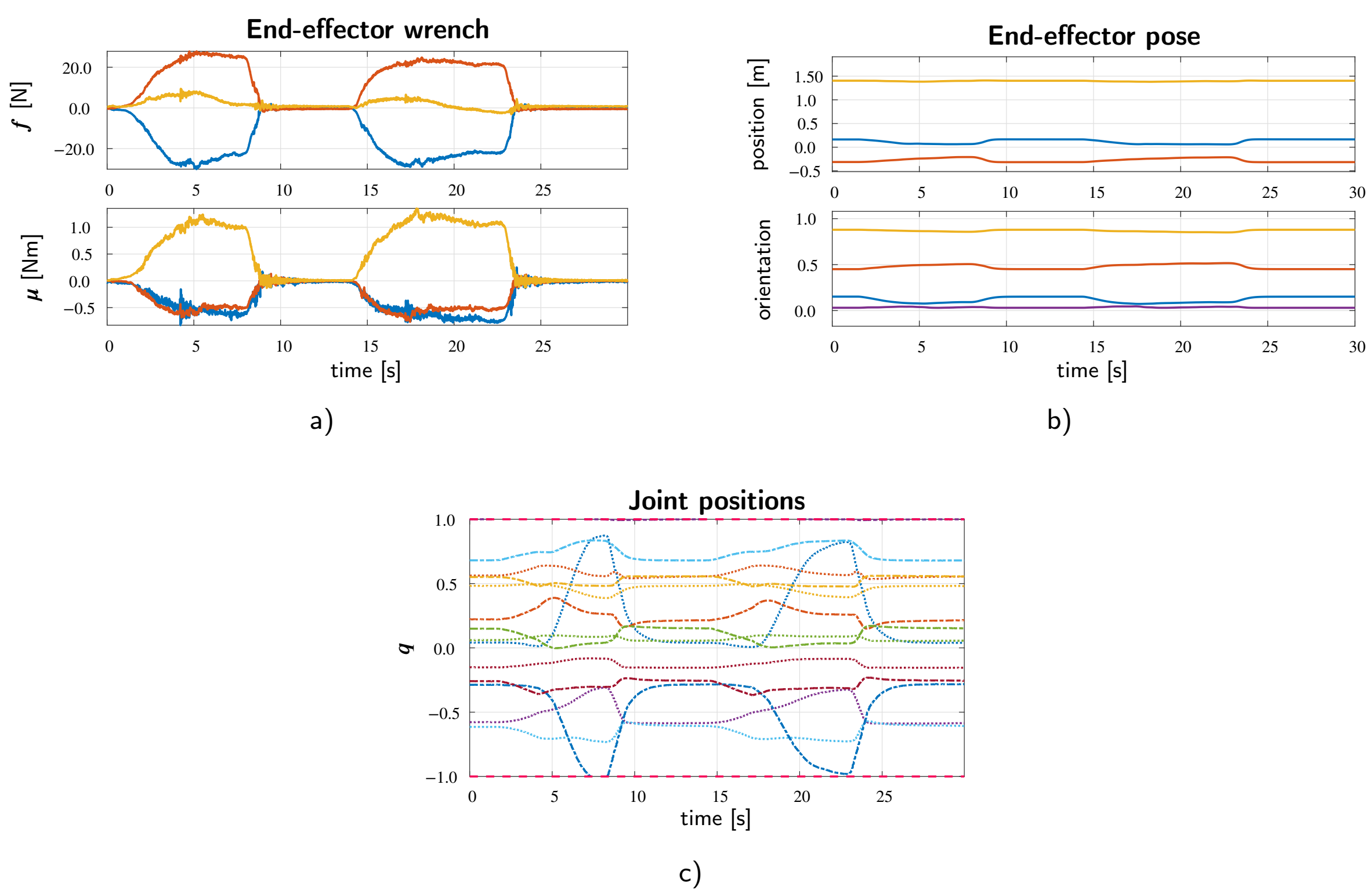

**Figure 13:** Optimization tasks validation including the optimization task in the hierarchy. a) Top: measured end-effector linear force. Bottom: measured end-effector moment. b) Top: end-effector position. Bottom: end-effector orientation expressed as quaternion. c) Normalized joint positions (dotted line for left arm, dot-dashed line for right arm).

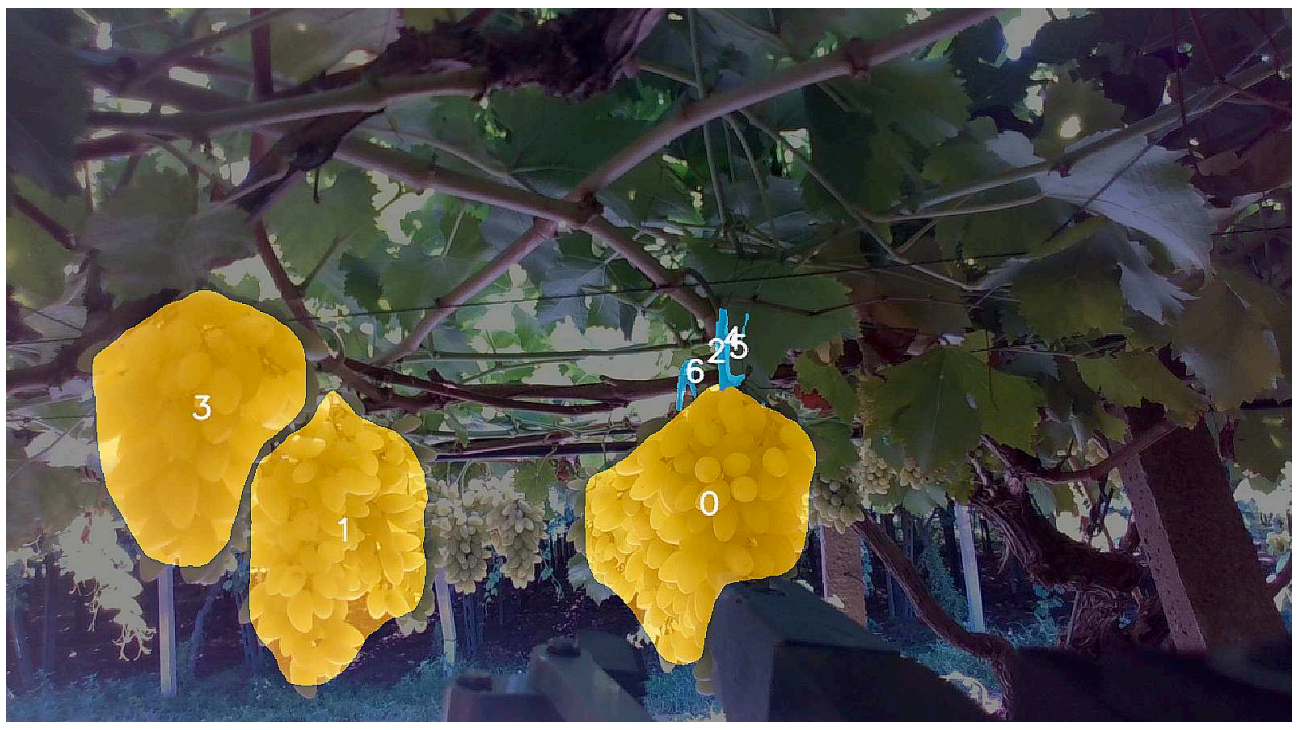

**Figure 14:** Output of the perception system. The highlighted zones are the grapes (yellow zones) and the peduncles (light-blue zones) that the software has detected and whose confidence score is above 0.9.

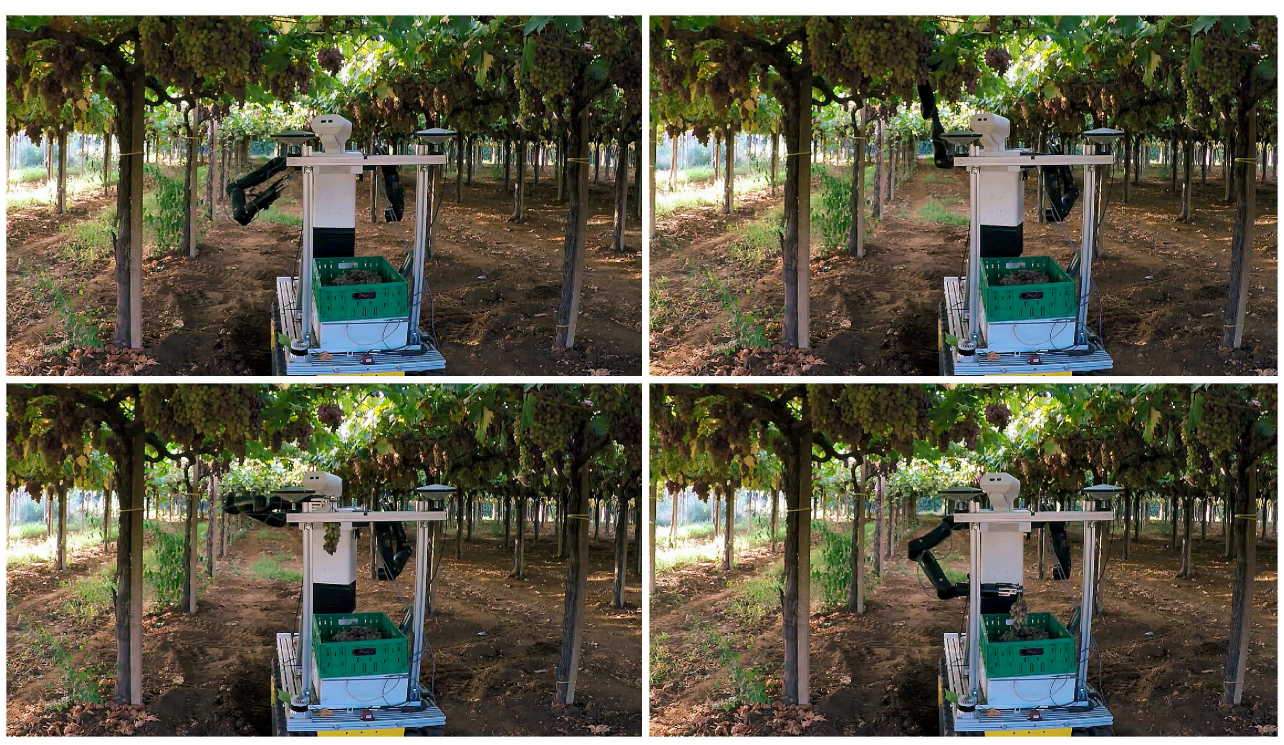

**Figure 15:** Top Left: image of the robot with both the harvesting tools in the home configuration. Top Right: image of the robot with the left harvesting tool in the so-called pre-release position. Bottom left: image of the robot with the left harvesting tool in the so-called release position.

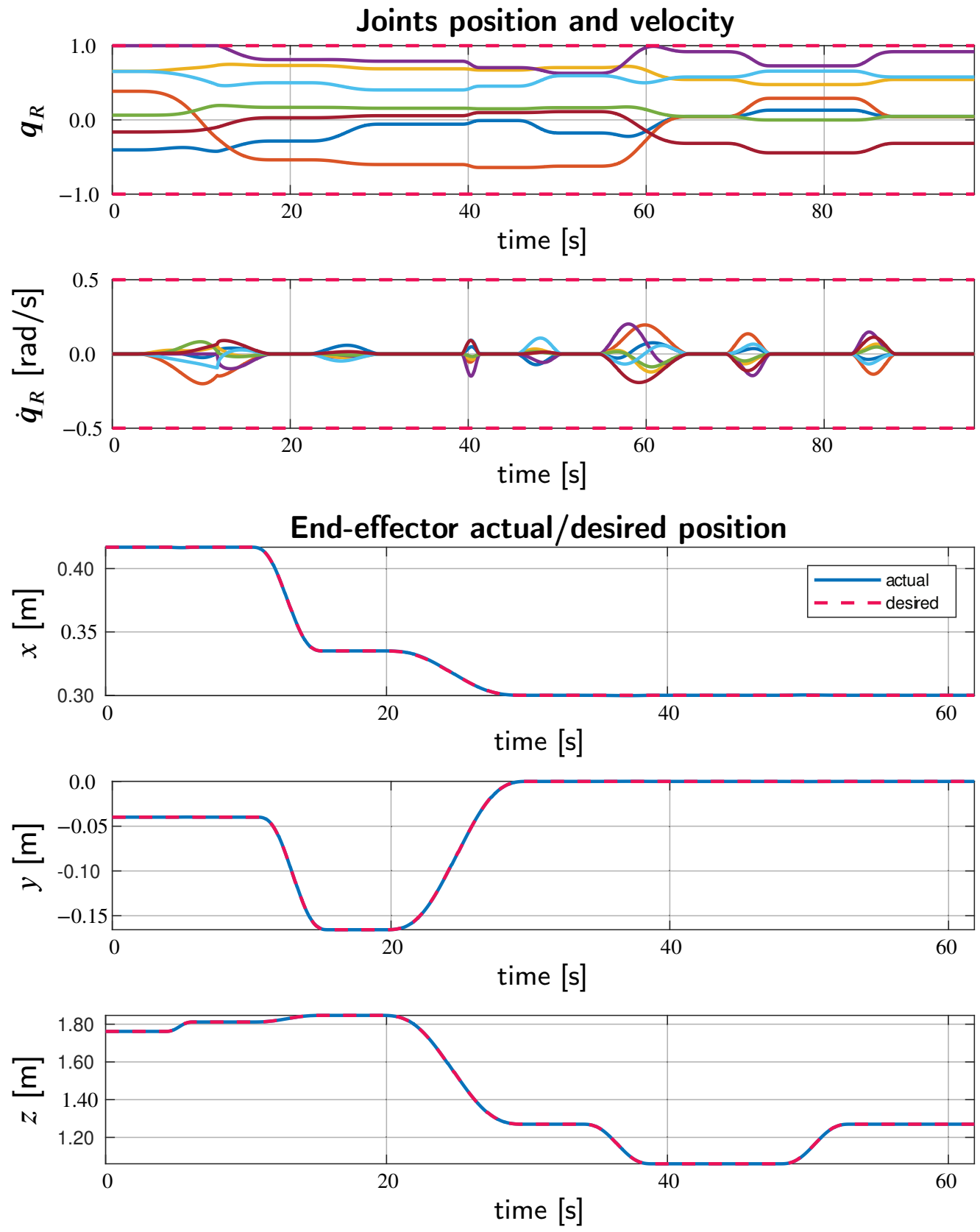

**Figure 16:** Top. Normalized joint positions and joint velocities and normalized lower and upper bounds for the right arm. Bottom. End-effector $x$, $y$ and $z$ actual (blue-solid line) and desired (red-dashed line) coordinates over time.

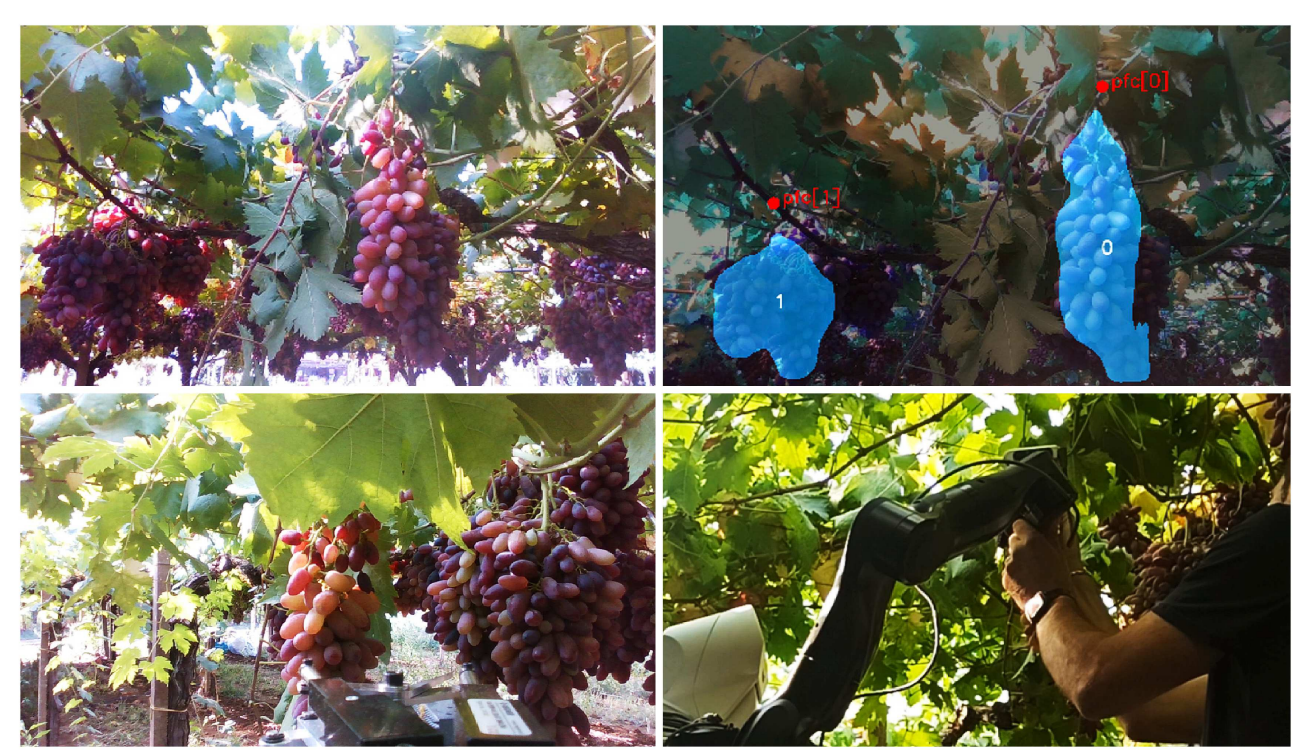

**Figure 17:** Top Left: image of the grapes from the head camera of the robot. Top Right: grapes recognition and peduncle localization. Bottom left: images of grapes taken from the right wrist camera. Bottom Right: human assisting the robot for grape harvesting.

## 7. Conclusions

This paper presents a control architecture designed to enable harvesting tasks in complex settings. The proposed architecture allows a bi-manual mobile robot to autonomously manage system safety and internal configurations while performing agronomic tasks and potentially interacting with a human operator. The elementary control objectives are detailed and integrated into the CBF and HQP frameworks. To provide an effective and straightforward solution, a finite-state machine algorithm was developed to switch operational tasks in the HQP hierarchy based on the perception system's performance, thereby adjusting the system's level of autonomy. Extensive experimental validation in both laboratory and real-field environments, conducted as part of the EU-funded project CANOPIES, demonstrated the flexibility and effectiveness of the proposed approach. Future efforts will focus on increasing the system's autonomy by incorporating LLM-based approaches, on adjusting the admittance parameters based on robot and human's parameters, conducting extensive field experiments including additional tasks such as pruning. Finally, the inclusion in the control architecture of more complex motion planners and perception algorithms to estimate and avoid external obstacles will be considered, being this an important aspect in the scenario under consideration.

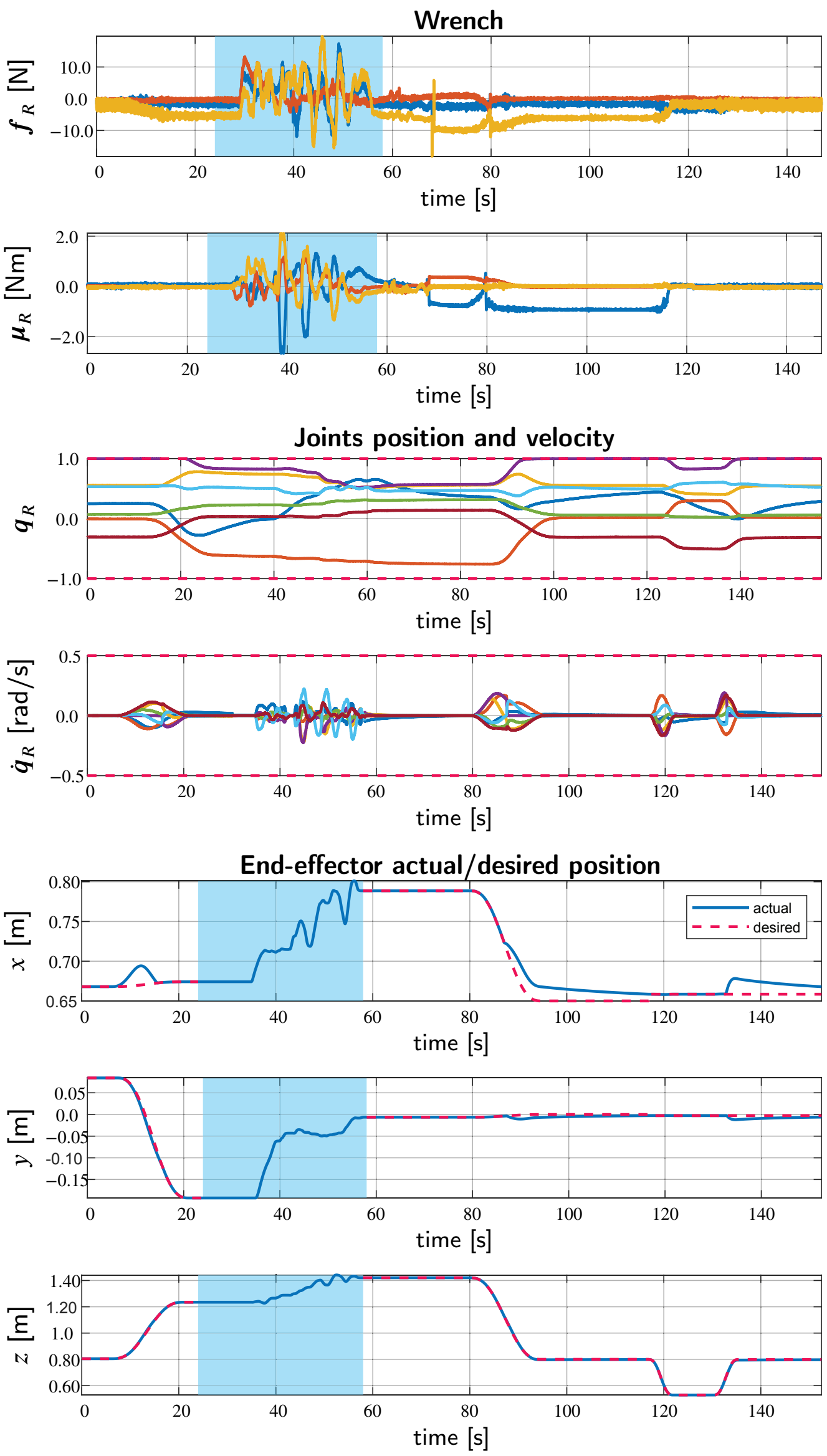

**Figure 18:** Top. Interaction wrench of the right end-effector over time during the harvesting experiment. Middle. Normalized joint positions and joint velocities and normalized lower and upper bounds for the right arm. Bottom. End-effector $x$, $y$ and $z$ actual (blue-solid line) and desired (red-dashed line) coordinates over time. The cyan background highlights the phase in which the system is in *guided* mode.

## Acknowledgments

This work was supported by H2020-ICT project CANOPIES-A Collaborative Paradigm for Human Workers and Multi-Robot Teams in Precision Agriculture Systems (Grant Agreement N. 101016906) and from Project CONCERTO-A COgNitive arChitecture for sEamless human-Robot inTeractiOn, CUP H53C24001050006, funded by EU in NextGenera- tionEU plan through the Italian "Bando Prin 2022 - D.D. 104 del 02-02-2022" by MUR.